\documentclass[authoryear]{elsarticle}

\usepackage{lineno,hyperref}
\usepackage{hyperref}
\usepackage{amssymb,amsmath,array,amsbsy}
\usepackage{amsthm}
\usepackage{caption}
\usepackage{subcaption}
\usepackage{microtype}

\modulolinenumbers[5]

\newtheorem{Remark}{Remark}

\begin{document}

\begin{frontmatter}


\title{Block modelling in dynamic networks with non homogeneous Poisson
  processes and exact ICL}


\author[mymainaddress]{Marco Corneli}
\author[mymainaddress]{Pierre Latouche}
\author[mymainaddress]{Fabrice Rossi}


\address[mymainaddress]{Universit\'e Paris 1 Panth\'eon-Sorbonne - Laboratoire SAMM \\
90 rue de Tolbiac, F-75634 Paris Cedex 13 - France}

\begin{abstract}
We develop a model in which interactions between nodes of a 
dynamic network are counted by non homogeneous Poisson processes.
In a block modelling perspective, nodes belong to hidden clusters (whose number is unknown) and the intensity functions of the counting processes only depend on the clusters of nodes.
In order to make inference tractable we move
to discrete time by partitioning
the entire time horizon in which interactions are observed
in fixed-length time sub-intervals.
First,  we  derive  an   exact  integrated  classification  likelihood
criterion  and maximize it relying on a greedy 
search approach. This allows to estimate the memberships to clusters
and the number of clusters simultaneously.
Then a maximum-likelihood estimator is
developed to estimate non parametrically
the integrated intensities.
We discuss the over-fitting problems 
of the model and propose a regularized
version solving these issues.
Experiments on real and simulated data
are carried out in order to assess
the proposed methodology.

\end{abstract}

\begin{keyword}
\texttt{Dynamic network \sep Stochastic block model\sep exact ICL \sep Non homogeneous Poisson Process.}
\end{keyword}

\end{frontmatter}

\linenumbers

\section{Introduction}
Graph clustering \citep{Schaeffer:COSREV2007} is probably one of the main
exploratory tools used in network analysis as it provides data analysts with a
high level summarized view of complex networks. One of the main paradigms for
graph clustering is community search \citep{FortunatoSurveyGraphs2010}: a
\emph{community} is a subset of nodes in a graph that are densely connected
and have relatively few connections to nodes outside of the community. While
this paradigm is very successful in many applications, it suffers from a
main limitation: it cannot be used to detect other important structures that
arise in graphs, such as bipartite structures, hubs, authorities, and other
patterns. 

The alternative solution favoured in this paper is provided by block models
\citep{White1971,White1976}: in such a model, a cluster consists of nodes that
share the same connectivity patterns to other clusters, regardless of the
pattern itself (community, hub, bipartite, etc.). A popular probabilistic view
on block models is provided by the stochastic block model
\citep[SBM,][]{Hollands83,wang1987}. The main idea is to assume that a hidden random variable
is attached to each node. This variable contains the cluster membership
information while connection probabilities between clusters are handled by the
parameters of the model. The reader is send to \cite{Goldenberg09} for a survey of probabilistic models for graphs and to \cite{wasserman1994social}, Ch.16, for an overview of the stochastic block models 

This paper focuses on dynamic graphs in the following sense: we assume that
nodes of the graph are fixed and that interactions between them are directed and
take place at a specific instant. In other words, we consider a directed
multi-graph (two nodes can be connected by more than one edge) in which each
directed edge is labelled with an occurrence time. We are interested in
extending the SBM to this type of graphs. More precisely, the proposed model is based on a counting process point of
view of the interactions between nodes: we assume that the number of
interactions between two nodes follows a non homogeneous Poisson counting
process (NHPP). As in a standard SBM, nodes are assumed to belong to clusters
that do not change over time, thus the temporal aspect is handled only via the
non homogeneity of the counting processes. Then the block model hypothesis
take the following form: the intensity of the NHPP that counts interactions
between two nodes depends only on the clusters of the nodes. In order to
obtain a tractable inference, a segmentation of the time interval under study
is introduced and the interactions are aggregated over the sub-intervals of
the partition. Following \cite{Come_Latouche15}, the model is adjusted to the
data via the maximization of the integrated classification likelihood
\citep[ICL][]{biernacki2000assessing} in an exact form. As in \cite{Come_Latouche15}
(and \cite{wyse2014inferring} for latent block models), the maximization is
done via a greedy search. This allows us to chose automatically the number of
clusters in the block model.

When the number of sub-intervals is large, the model can suffer from a form of over fitting   as  the   ICL   penalizes  only   a   large  number   of
clusters. Therefore, we introduce
a variant, based on the model developed in \cite{corneli_asonam}, in which sub-intervals are clustered into classes of
homogeneous intensities. Those clusters are accounted for in a new version of the ICL which prevents over fitting. 

This paper is structured as
follows: in Section \ref{sec:RW} we mention works related to the approach we propose, Section \ref{sec:TheMod} presents the proposed temporal extension of
the SBM, Section \ref{sec:Est} derives the exact ICL for this model and presents the greedy search algorithm used to maximize
the ICL. Section \ref{sec:Exp} gathers experimental results on
simulated data and on real world data.  Section \ref{sec:Conc} concludes the paper.

\section{Related Works}
\label{sec:RW}
Numerous extensions of the original SBM have already been proposed to deal
with dynamic graphs. In this context, both nodes memberships to a cluster and
interactions between nodes can be seen as stochastic processes.  In
\cite{yang2011detecting}, for instance, authors introduce a Markov Chain to
obtain the cluster of node in $t$ given its cluster at time $t-1$.  \cite{xu2013dynamic} as well as \cite{xing2010} used a state space model to describe temporal changes at the
level of the connectivity pattern. In the latter, the authors developed a
method to retrieve overlapping clusters through time. In general, the proposed
temporal variations of the SBM share a similar approach: the data set consists
in a sequence of graphs rather than the more general structure we assume.
Some papers remove those assumptions by considering continuous time models in
which edges occur at specific instants (for instance when someone sends an
email). This is the case of e.g. \cite{proceedingsdubois2013} and of
\cite{Rossi12,GuigouresEtAl2015}. 
A temporal stochastic block model, related to the one presented in this paper is independently developed by \cite{Matias_Poisson}. They assume that nodes in a network belong to clusters whose composition do not change over time and interactions are counted by a non-homogeneous Poisson process whose intensity only depends on the nodes clusters. In order to estimate (non-parametrically) the instantaneous intensity functions of the Poisson processes, they develop a variational EM algorithm to maximize an approximation of the likelihood.

\section{The model}\label{sec:TheMod}
We consider a fixed set of $N$ nodes, $\{1,\ldots,N\}$, that can interact as
frequently as wanted during the time interval $[0,T]$. Interactions are
directed from one node to another and are assumed to be
instantaneous\footnote{In practice, the starting time of an interaction with a
  duration will be considered.}. A natural mathematical model for this type of
interactions is provided by counting processes on $[0,T]$. Indeed a counting
process is a stochastic process with values that are non negative integers
increasing through time: the value at time $t$ can be seen as the number of
interactions that took place from $0$ to $t$. Then the classical adjacency
matrix $(X_{ij})_{1\leq i, j\leq N}$ of static graphs is replaced by a
$N\times N$ collection of counting processes,
$(X_{ij}(t))_{1\leq i, j\leq N}$, where $X_{ij}(t)$ is the counting process
that gives the number of interactions from node $i$ to node $j$. We still call
$\mathbf{X}=(X_{ij}(t))_{1\leq i, j\leq N}$ the adjacency matrix of this dynamical graph.

We introduce in this Section a generative model for adjacency matrices of
dynamical graphs that is inspired by the classical stochastic block model
(SBM). 

\subsection{Non-homogeneous Poisson counting process}
We first chose a simple form for $X_{ij}(t)$: we assume that this process is a
non-homogeneous Poisson counting process (NHPP) with instantaneous intensity
given by the function from $[0,T]$ to $\mathbb{R}$, $\lambda_{ij}$. For
$s \leq t \leq T$, it then holds
\begin{equation}
 p(X_{ij}(t) - X_{ij}(s)|\lambda_{ij})=\frac{(\int_s^t \lambda_{ij}(u)du)^{X_{ij}(t) - X_{ij}(s)}}{(X_{ij}(t) - X_{ij}(s))!}\exp\left(-\int_s^t \lambda_{ij}(u)du\right), 
\end{equation}
where $X_{ij}(t) - X_{ij}(s)$ is the (non negative) number of
interactions from $i$ to $j$ that took place during $[s,t]$. (We assume that
$X_{ij}(0)=0$.)

\subsection{Block modelling}
The main idea of the SBM \citep{Hollands83, wang1987} is to assume that nodes
have some (hidden) characteristics that solely explain their interactions, in
a stochastic sense. In our context this means that rather than having pairwise
intensity functions $\lambda_{ij}$, those functions are shared by nodes that
have the same characteristics. 

In more technical terms, we assume the nodes are grouped in $K$ clusters
($\mathcal{A}_1,\dots, \mathcal{A}_K$) and introduce a hidden cluster membership random vector
$\mathbf{z}\in \{1,\ldots K\}^N$ such that
\begin{equation*}
 z_i=k \qquad\text{iff}\qquad i \in \mathcal{A}_k, \qquad k \leq K. 
\end{equation*}
The random component $z_i$ is assumed to follow a multinomial distribution
with parameter vector $\omega$ such that 
\begin{equation*}
  \mathbb{P}\{ z_i=k \}=\omega_k \qquad\text{with}\qquad \sum_{k \leq K} \omega_k=1.
\end{equation*}
In addition, the $(z_i)_{1\leq i\leq N}$ are assumed to be independent
(knowing $\omega$) and thus
 \begin{equation}\label{eq:z:omega}
  p(\mathbf{z} | \boldsymbol{\omega},K)=\prod_{k \leq K}\omega_k^{|\mathcal{A}_k|},
 \end{equation}
where $|\mathcal{A}_k|$ denotes the cardinal of $\mathcal{A}_k$. Notice that this part of the
model is exactly identical to what is done in the classical SBM. 

In a second step, we assume that given $\mathbf{z}$, the counting processes
$X_{ij}(t)$ are independent and in addition that the intensity function
$\lambda_{ij}$ depends only on $z_i$ and $z_j$. In order to keep notations
tight we denote $\lambda_{z_iz_j}$ the common intensity function and we will
not use directly the pairwise intensity functions $\lambda_{ij}$. We denote
$\boldsymbol{\lambda}$ the matrix valued intensity function
$\boldsymbol{\lambda}=(\lambda_{kg}(t))_{1\leq k,g \leq  K}$. 

Combining all the assumptions, we have for $s \leq t \leq T$
\begin{equation}
  \label{eq:fullmodel}
p(\mathbf{X}(t)-\mathbf{X}(s)|\mathbf{z},\boldsymbol{\lambda})=\prod_{i\neq j}  \frac{(\int_s^t \lambda_{z_iz_j}(u)du)^{X_{ij}(t)-X_{ij}(s)}}{(X_{ij}(t) - X_{ij}(s))!}\exp\left(-\int_s^t \lambda_{z_iz_j}(u)du\right).
\end{equation}

\subsection{Discrete time version}
In order to make inference tractable, we move from the continuous time model
to a discrete time one. This is done via a partition of the interval $[0,T]$
based on a set of $U+1$ instants
\begin{equation*}
 0=t_0 \leq t_1 \leq \dots \leq t_{U-1} \leq t_U=T,
\end{equation*}
that defines $U$ intervals $I_u:=[t_{u-1}, t_u[$ (with arbitrary length
$\Delta_u$).
The purpose of the partition is to aggregate the interaction. Let us denote
\begin{equation}
Y_{ij}^{I_u}:=X_{ij}(t_u) - X_{ij}(t_{u-1}), \qquad u \in \{1, \dots, U \}.
\end{equation}
In words, $Y_{ij}^{I_u}$ measures the increment, over the time interval $I_u$,
of the Poisson process counting interactions from $i$ to $j$. We denote by $Y_{ij}$ the random vector
\begin{equation*}
 Y_{ij}:=(Y_{ij}^{I_1}, \dots , Y_{ij}^{I_U})^T.
\end{equation*}
Thanks to the independence of the increments of a Poisson process, we get the following joint density:
\begin{equation}
\label{eq:L1}
p(Y_{ij} | \lambda_{ij})= \prod_{u=1}^{U} \left(\frac{(\int_{I_u} \lambda_{ij}(s)ds)^{Y_{ij}^{I_u}}}{Y_{ij}^{I_u}!}\exp{\left(-\int_{I_u} \lambda_{ij}(s)ds\right)}\right).
\end{equation} 
The variations of $\lambda_{ij}$ inside an interval $I_u$ have no effect on the
distribution of $Y_{ij}$. This allows us to use the integrated intensity
function $\Lambda$ defined on $[0,T]$ by
\begin{equation*}
 \Lambda_{ij}(t) :=\int_{0}^t \lambda_{ij}(s)ds.
\end{equation*}
In addition, we denote by $\pi_{ij}^{I_u}$ the increment of the integrated
intensity function over $I_u$ 
\begin{equation*}
\pi_{ij}^{I_u}:=\Lambda_{ij}(t_u)- \Lambda_{ij}(t_{u-1}), \qquad\forall u \in \{1,\dots, U\}.
\end{equation*}
Then equation \eqref{eq:L1} becomes
\begin{equation}
 \label{eq:L2}
 p(Y_{ij} | \pi_{ij})= \prod_{u=1}^{U} \left(\frac{(\pi_{ij}^{I_u})^{Y_{ij}^{I_u}}}{Y_{ij}^{I_u}!}\exp{\left( -\pi_{ij}^{I_u}\right)}\right),
\end{equation}
with $\pi_{ij}:=(\pi_{ij}^{I_1}, \dots, \pi_{ij}^{I_U})^T$.

Using the block model assumptions, we have in addition
\begin{equation}
  \label{eq:Y:bm}
 p(Y_{ij} | \pi_{z_iz_j},z_i,z_j)= \prod_{u=1}^{U} \left(\frac{(\pi_{z_iz_j}^{I_u})^{Y_{ij}^{I_u}}}{Y_{ij}^{I_u}!}\exp{\left( -\pi_{z_iz_j}^{I_u}\right)}\right),
\end{equation}
where we have used the fact that $\lambda_{ij}=\lambda_{z_iz_j}$ (which leads
to $\Lambda_{ij}=\Lambda_{z_iz_j}$, etc.).  

Considering the network as a whole, we can introduce two tensors of order
3. $Y$ is a $N\times N\times U$ random tensor whose element $(i,j,u)$ is the
random variable $Y_{ij}^{I_u}$ and $\pi$ is the $K \times K \times U$ tensor
whose element $(k,g,u)$ is $\pi_{kg}^{I_u}$. $Y$ can be seen as an aggregated
(or discrete time version) of the adjacency process $\mathbf{X}$ while $\pi$
can be seen as summary of the matrix valued intensity function
$\boldsymbol{\lambda}$. 

The conditional independence assumption of the block model leads to
\begin{equation}
 \label{eq:L4}
 p(Y|\pi,\mathbf{z})=\prod_{i,j}^{N}p(Y_{ij}|\pi_{z_iz_j},z_i,z_j). 
\end{equation}   
To simplify the rest of the paper, we will use the following notations
\begin{align*}
\prod_{i,j} \prod_{k,g}\prod_{u}&:=\prod_{i=1}^N \prod_{j=1}^N \prod_{k=1}^K \prod_{g=1}^K \prod_{u=1}^U\\
\prod_{z_i=k} \left(\prod _{z_j=g}\right)&:=\prod_{\substack{i: \\ z_i=k}} \left(\prod_{\substack{j: \\ z_j=g}}\right).
\end{align*}
The joint distribution of $Y$, given $\textbf{z}$ and $\pi$, is
\begin{align}
\label{eq:L5}
p(Y|\mathbf{z}, \pi) &= \prod_{i,j} \prod_u \left(\frac{(\pi_{z_i z_j}^{I_u})^{Y_{ij}^{I_u}}}      {Y_{ij}^{I_u}!}\exp{\left( -\pi_{z_i z_j}^{I_u}\right)}\right) \nonumber \\
&= \prod_{k,g} \prod_{u}\left(\frac{(\pi_{k,g}^{I_u})^{S_{kgu}}}      {P_{kgu}}\exp{\left( -|A_k||A_g|\pi_{kg}^{I_u}\right)}\right),
\end{align} 
where
\begin{equation*}
 S_{kgu}=\sum_{\substack{ z_i=k}}\sum_{\substack{ z_j=g}} Y_{ij}^{I_u}, 
\end{equation*}
is the total number of interactions from cluster $k$ to cluster $g$ (possibly
equal to $k$) and with
\begin{equation*}
P_{kgu}=\prod_{\substack{z_i=k}}\prod_{z_j=g} Y_{ij}^{I_u}!.
\end{equation*}

\subsection{A constrained version}
As will be shown in Section \ref{par:GS}, the model presented thus far is
prone to over fitting when the number of sub-intervals $U$ is large compared
to $N$. Additional constraints on the intensity functions
$\{\Lambda_{kg}(t)\}_{k,g \leq K}$ are needed in this situation. 

Let us consider a fixed pair of clusters $(k,g)$. So far, the increments
$\{\pi_{kg}^{I_u}\}_{u \leq U}$ are allowed to differ on each $I_u$ over the
considered partition. A constraint can be introduced by assigning the time
intervals $(I_1,\dots I_U)$ to different time clusters and assuming that
increments are identical for all the intervals belonging to the same time
cluster. Formally, we introduce $D$ clusters ($\mathcal{C}_1,\dots, \mathcal{C}_D$) and a hidden
random vector $\mathbf{y}\in \{0,1\}^U$, labelling memberships
\begin{equation*}
 y_u=d \qquad\text{iff}\qquad I_u \in \mathcal{C}_d.
\end{equation*}
Each $y_u$ is assume to follow a multinomial distribution depending on parameter $\boldsymbol{\rho}$
\begin{equation*}
  \mathbb{P}\{ y_u=d \}=\rho_d \qquad\text{with}\qquad \sum_{d \leq D} \rho_d=1,
\end{equation*}
and in addition the $y_u$ are assumed to be independent, leading to
 \begin{equation}\label{eq:y:rho}
  p(\mathbf{y} | \boldsymbol{\rho},D)=\prod_{d \leq D}\rho_d^{|\mathcal{C}_d|}.
 \end{equation}
The random variable $Y_{ij}^{I_u}$ is now assumed to follow the conditional distribution
\begin{equation}
 p(Y_{ij}^{I_u}| \mathbf{z}, \mathbf{y})=\frac{(\pi_{z_i z_j}^{y_u})^{Y_{ij}^{I_u}}}{Y_{ij}^{I_u}!}\exp{(-\pi_{z_i z_j}^{y_u})}.
\end{equation}
Notice that the new Poisson parameter $\pi_{z_i z_j}^{y_u}$ replaces
$\pi_{z_i z_j}^{I_u}$ in the unconstrained version.  The joint distribution of
$Y$, given $\mathbf{z}$ and $\mathbf{y}$, can easily be
obtained 
\begin{equation}
\label{eq:L_unconstrained}
p(Y|\mathbf{z}, \mathbf{y}, \pi)=\prod_{k,g} \prod_{d}\left(\frac{(\pi_{kg}^d)^{S_{kgd}}}{P_{kgd}}\exp{\left( -|\mathcal{A}_k||\mathcal{A}_g||\mathcal{C}_d|\pi_{kg}^{d}\right)}\right),
\end{equation}
where
\begin{equation*}
S_{kgd}=\sum_{\substack{z_i=k}}\sum_{\substack{z_j=g}}\sum_{y_u=d}
Y_{ij}^{I_u}, \qquad
P_{kgd}=\prod_{\substack{z_i=k}}\prod_{z_j=g}\prod_{y_u=d} Y_{ij}^{I_u}!.
\end{equation*}
\begin{Remark}
  The introduction of this hidden vector $\mathbf{y}$ is not the only way to
  impose regularity constraints to the integrated function
  $\Lambda_{kg}(t)$. For example, a segmentation constraint could be imposed
  by forcing each temporal cluster to contain only adjacent time intervals.
\end{Remark}

\subsubsection{Summary}
We have defined two generative models:
\begin{description}
\item[Model A] the model has two meta parameters, $K$ the number of clusters
  and $\boldsymbol{\omega}$ the parameters of a multinomial distribution on
  $\{1,\ldots,K\}$. The hidden variable $\mathbf{z}$ is generated by the
  multivariate multinomial distribution of equation \eqref{eq:z:omega}. Then
  the model has a $K\times K\times U$ tensor of parameters $\pi$. Given
  $\mathbf{z}$ and $\pi$, the model generates a tensor of interaction counts
  $Y$ using equation \eqref{eq:L5}. 
\item[Model B] is a constrained version of model \textbf{A}. In addition to the meta
  parameters $K$ and $\boldsymbol{\omega}$ of model \textbf{A}, it has two meta parameters,
  $D$ the number of clusters of time sub-intervals and $\boldsymbol{\rho}$ the
  parameters of a multinomial distribution on
  $\{1,\ldots,D\}$. The hidden variable $\mathbf{y}$ is generated by the
  multivariate multinomial distribution of equation \eqref{eq:y:rho}. Model
  \textbf{B} has a $K\times K\times D$ tensor of parameters $\pi$. Given
  $\mathbf{z}$, $\mathbf{y}$ and $\pi$, the model generates a tensor of interaction counts
  $Y$ using equation \eqref{eq:L_unconstrained}. 
\end{description}
Unless specified otherwise ``the model'' is used for model \textbf{A}. 

\section{Estimation}\label{sec:Est}
\subsection{Non parametric estimation of integrated intensities}
In this Section we assume that $\mathbf{z}$ is known. No hypothesis has been formulated about the shape of the functions $\{\Lambda_{kg}(t)\}_{\{k,g \leq K, t \leq T\}}$ and the increments of these functions over the partition introduced can be estimated by maximum likelihood (ML), thanks to equation \eqref{eq:L5}
\begin{equation*}
 \log\mathcal{L}(\pi| Y, \mathbf{z})=\sum_{k,g}\sum_{u} \left[S_{kgu}\log(\pi_{kg}^{I_u}) - |\mathcal{A}_k||\mathcal{A}_g|\pi_{kg}^{I_u} + c \right],
\end{equation*}
where $c$ denotes those terms not depending on $\pi$. It immediately follows
\begin{equation}
\label{eq:MLE_pi}
  \hat{\pi}_{kg}^{I_u}=\frac{S_{kgu}}{|\mathcal{A}_k||\mathcal{A}_g|}, \qquad\forall (k,g),
\end{equation}
where $\hat{\pi}_{kg}^{I_u}$ denotes the ML estimator of $\pi_{kg}^{I_u}$. 
In words, $\Lambda_{kg}(t_u) - \Lambda_{kg}(t_{u-1})$ can be estimated by ML as the total number of interactions on the sub-graph corresponding to the connections from cluster $A_k$ to cluster $A_g$, over the time interval $I_u$, divided by the number of nodes on this sub-graph. Once the tensor $\pi$ has been estimated, we have a point-wise, non parametric estimator of $\Lambda_{kg}(t_u)$, for every $u \leq U$, defined by
\begin{equation}
\label{eq:MLE_Lambda}
 \hat{\Lambda}_{kg}(t_u)= \sum_{l=1}^u \hat{\pi}_{kg}^{I_l}, \qquad\forall (k,g).
\end{equation}
Thanks to the properties of the ML estimator, together with the linearity of \eqref{eq:MLE_Lambda}, we know that $\hat{\Lambda}_{kg}(t_u)$ is an unbiased and convergent estimator of $\Lambda_{kg}(t_u)$.

\begin{Remark}
Estimator \eqref{eq:MLE_Lambda} at times $\{t_u\}_{u\leq U}$, can be viewed as an extension to random graphs and mixture models of the non parametric estimator proposed in \cite{Leemis91}.  In that article, $N$-trajectories of independent NHPPs, sharing the same intensity function, are observed and the proposed estimator is basically obtained via method of moments.
\end{Remark}   

In all the experiments, we consider the following step-wise linear estimator of $\Lambda_{kg}(t)$
\begin{equation}
\label{eq:linear_estimator}
 \hat{\Lambda}_{kg}(t)=\sum_{u=1}^{U} \left[\hat{\Lambda}_{kg}(t_{u-1}) + \frac{\hat{\Lambda}_{kg}(t_u) - \hat{\Lambda}_{kg}(t_{u-1})}{t_u - t_{u-1}}(t - t_{u-1})\right]\mathbf{1}_{[t_{u-1}, t_u[}(t),
\end{equation}
which is a linear combination of estimators in equation \eqref{eq:MLE_Lambda} on the interval $[0,T]$. This is a consistent and unbiased estimator of $\Lambda_{kg}(t)$ at times $\{t_u\}_{u \leq U}$ only. 

When considering model \textbf{B}, equations \eqref{eq:MLE_pi} and \eqref{eq:MLE_Lambda} are replaced by
\begin{align}
\label{eq:MLE_modelB}
  &\hat{\pi}_{kg}^{d}=\frac{S_{kgd}}{|\mathcal{A}_k||\mathcal{A}_g||\mathcal{C}_d|} \\
  &\hat{\Lambda}_{kg}(t_u)= \sum_{l=1}^u \hat{\pi}_{kg}^{y_l}. 	
\end{align}
Equation \eqref{eq:linear_estimator} remains unchanged, but an important difference between the constrained model and the unconstrained one should be understood: in the former, each interval $I_u$ corresponds to a different slope for the function $\hat{\Lambda}_{kg}(t)$ whereas in the latter we only have $D$ different slopes, one for each time cluster. 

\subsection{ICL}
Since the vector $\mathbf{z}$, as well as the number of clusters $K$ are unknown, estimator \eqref{eq:MLE_pi} cannot be used directly. Hence we propose a two step procedure consisting in
\begin{enumerate}
\item providing estimates of $\mathbf{z}$ and $K$,
\item using these estimates to implement \eqref{eq:MLE_pi} and \eqref{eq:MLE_Lambda}.
\end{enumerate}
To accomplish the first task, the same approach followed in \cite{Come_Latouche15} is adopted: we directly maximize the the joint integrated log-likelihood of complete data (ICL), relying on a greedy search over the labels and number of clusters. To perform such a maximization, we need the ICL to have an explicit form. This can be achieved by introducing conjugated prior distributions on the model parameters.
The ICL can be written as
\begin{equation}
\label{eq:ICL}
\mathcal{ICL}(\mathbf{z}, K):=\log(p(Y,\mathbf{z}|K))=\log(p(Y|\mathbf{z}, K)) + \log(p(\mathbf{z}|K)).   
\end{equation}
This \emph{exact} quantity is approximated by the well known ICL \emph{criterion} \citep{biernacki2000assessing}. This criterion, obtained through Laplace and Stirling approximations of the joint density on the left hand side of equation \eqref{eq:ICL}, is used as a model selection tool, since it penalizes models with a high number of parameters. In the following, we refer to the joint log-density in equation \eqref{eq:ICL} as to the \emph{exact ICL} to differentiate it from the ICL criterion.

We are now going to study in detail the two quantities on the r.h.s. of the above equation.
The first probability density is obtained by integrating out the parameter $\pi$
\begin{equation*}
p(Y|\mathbf{z}, K)= \int p(Y,\pi|\mathbf{z},K)d\pi.
\end{equation*}
In order to have an explicit formula for this term, we impose the following Gamma prior conjugated density over the tensor $\pi$:
\begin{equation*}
p(\pi|a,b)=\prod_{k,g,u}\frac{b^a}{\Gamma(a)}\pi_{kgu}^{a-1}e^{-b \pi_{kgu}},
\end{equation*}
where the hyper-parameters of the Gamma prior distribution have been set constant to $a$ and $b$ for simplicity.\footnote{The model can easily be extended to the more general framework:
\begin{equation*}
 p(\pi_{kgu}|a_{kgu}, b_{kgu})=\text{Gamma}(\pi_{kgu}|a_{kgu}, b_{kgu}).
\end{equation*} 
}
By using the Bayes rule
\begin{equation*}
 p(Y,\pi|\mathbf{z})=p(Y|\pi,\mathbf{z})p(\pi|a,b),
\end{equation*}
we get:
\begin{align*}
 \begin{split}
 p(Y, \pi|\mathbf{z})= &\prod_{k,g,u} \frac{b^a}{\Gamma(a)P_{kgu}}\pi_{kgu}^{S_{kgu}+a-1}\\
 &\times\exp\left(-\pi_{kgu}\left[|\mathcal{A}_k||\mathcal{A}_g|+ b\right]\right),
\end{split}
\end{align*}
which can be integrated with respect to $\pi$ to obtain
\begin{align}
 \label{eq:T1}
\begin{split}
 p(Y|\mathbf{z},K)=&\prod_{k,g,u} \left[\frac{b^a}{\Gamma(a)P_{kgu}} \frac{\Gamma[S_{kgu}+a]}{\left[|\mathcal{A}_k||\mathcal{A}_g| + b\right]^{(S_{kgu}+a)}}\right].
\end{split}
\end{align}
We now focus on the second density on the right hand side
\begin{equation*}
 p(\mathbf{z}|K)=\int p(\mathbf{z},\boldsymbol{\omega}|K)d\boldsymbol{\omega}.
\end{equation*} 
A Dirichlet \emph{prior} distribution can be attached to $\boldsymbol{w}$ in order to get an explicit formula, in a similar fashion of what we did with $\pi$:
\begin{align*}
 \nu(\omega|K) =& \text{Dir}_K(\boldsymbol{\omega}; \alpha,\dots,\alpha).
\end{align*}
The integrated density $p(\mathbf{z}|K)$ can be proven to reduce to
\begin{align}
\label{eq:p_z}
 p(\mathbf{z}| K)=\frac{\Gamma(\alpha K)}{\Gamma(\alpha)^K}\frac{\prod_{k\leq K}\Gamma(|\mathcal{A}_k| + \alpha)}{ \Gamma(N + \alpha K)}
 \end{align}

\subsection{Model B}
When considering the constrained framework described at the end of the previous section, the ICL is defined
\begin{align*}
 \mathcal{ICL}(\mathbf{z}, \mathbf{y}, K,D):=&\log(p(Y,\mathbf{z}, \mathbf{y}|K,D)) \\
 =&\log(p(Y|\mathbf{z},\mathbf{y})) + \log(p(\mathbf{z}|K)) + \log(p(\mathbf{y}|D))
\end{align*}
and it is maximized to provide estimates of $\mathbf{z}, \mathbf{y}, K$ and $D$.
The first density on the right hand side is obtained by integrating out the hyper-parameter $\pi$. This integration can be done explicitly by attaching to $\pi$ the following prior density function
\begin{equation*}
 \nu(\pi|a,b)=\prod_{k,g}\prod_d \frac{b^a}{\Gamma(a)}\pi_{kgd}^{a-1}e^{-b \pi_{kgd}}.
\end{equation*}
The second integrated density on the right hand side can be read in \eqref{eq:p_z} and the third is obtained by integrating out the parameter $\boldsymbol{\rho}$, whose prior density density function is assumed to be 

\begin{equation*}
 \nu(\boldsymbol{\rho}|D)=\text{Dir}_D(\boldsymbol{\rho};\beta, \dots, \beta).
\end{equation*}
The exact ICL is finally obtained by taking the logarithm of
\begin{align}
\label{eq:modelB_ICL}
  p(Y,\mathbf{z}, \mathbf{y}|K,D)&=\prod_{k,g,d} \frac{b^a}{\Gamma(a)P_{kgd}} \frac{\Gamma[S_{kgd}+a]}{\left[|\mathcal{A}_k||\mathcal{A}_g||\mathcal{C}_d| + b\right]^{(S_{kgd}+a)}} \nonumber \\
  &\times \frac{\Gamma(\alpha K)}{\Gamma(\alpha)^K}\frac{\prod_{k\leq K}\Gamma(|\mathcal{A}_k| + \alpha)}{ \Gamma(N + \alpha K)}\nonumber \\
  &\times \frac{\Gamma(\beta D)}{\Gamma(\beta)^D}\frac{\prod_{d\leq D}\Gamma(|\mathcal{C}_d| + \beta)}{ \Gamma(U + \beta D)}.
\end{align}

\subsection{Greedy search}
\label{par:GS}
 By setting conjugated prior distributions over the model parameters, we obtained an ICL (equation \eqref{eq:ICL}) in an explicit form. Nonetheless explicit formulas to maximize it, with respect to $\mathbf{z}$ and $K$, do not exist. We then rely on a greedy search algorithm, that has been used to maximize the exact ICL, in the context of a standard SBM, by \cite{Come_Latouche15}.
This algorithm basically works as follows:
\begin{enumerate}
\item An initial configuration for both $\mathbf{z}$ and $K$ is set (standard clustering algorithms like \emph{k-means} or hierarchical clustering can be used).
\item Labels switches leading to the highest increase in the exact ICL are repeatedly made. A label switch consists in a merge of two clusters or in a node switch from one cluster to another. 
\end{enumerate}

\begin{Remark}
The greedy algorithm described in this section, makes the best choice \emph{locally}. A convergence toward the global optimum in not guaranteed and often this optimum can only be approximated by a local optimum reached by the algorithm.
\end{Remark}

\begin{Remark}
The \emph{exact ICL} (as well as the \emph{ICL criterion}) penalizes the number of parameters. Since the tensor $\pi$ has dimension $K \times K \times U$,  when $U$, which is fixed, is very hight, the ICL will take its maximum for $K=1$. In other words the only way the ICL has to make the model more parsimonious is to reduce $K$ up to one. By doing so, any community (or other) structure will not be detected. This over-fitting problem has nothing to see with the possible limitations of the greedy search algorithm and it can be solved by switching to model \textbf{B}. 
\end{Remark}

Once $K_{max}$ has been fixed, together with an initial value of $\mathbf{z}$, a shuffled sequence of all the nodes in the graph is created. Each node in the sequence is moved to the cluster leading to the highest increase in the ICL, if any. This procedure is repeated until no further increase in the ICL is still possible. Henceforth, we refer to this step as to \emph{Greedy-Exchange} (\textbf{GE}). 
When maximizing the modularity score to detect communities, the \textbf{GE} usually is a final refinement step to be adopted after repeatedly merging clusters of nodes. In that context, moreover, the number of clusters is initialized to $U$ and each node is alone in its own cluster. See for example \cite{NoakRotta}.  
Here, we follow a different approach, proposed by \cite{Come_Latouche15} and \cite{Blondel08fastunfolding}: after running the \text{GE} , we try to \emph{merge} the remaining clusters of nodes in the attempt to increase the ICL. In this final step (henceforth \textbf{GM}), all the possible merges are tested and the best one is retained.

The ICL does not have to be computed before and after each swap/merge: possible increases can be assessed directly. 
When switching one node (say $i$) from cluster $\mathcal{A}_{k'}$ to $\mathcal{A}_l$, with $k' \neq l$, the change in the ICL is given by\footnote{Hereafter, the ``*'' notation refers to the statistics \emph{after} switching/merging. 
}
 \begin{equation*}
  \Delta_{k' \rightarrow l} = ICL(\mathbf{z^*},K)- ICL(\mathbf{z},K).
 \end{equation*}
 The only statistics not simplifying, are those involving  $k'$ and $l$, hence the equation above can be read as follows
 \small
 \begin{align}
 \label{eq:switch}
 \begin{split}
 \Delta_{k'\rightarrow l} :=& \log\left(\frac{\Gamma(|\mathcal{A}_{k'}| - 1 + \alpha)\Gamma(|\mathcal{A}_l|+1 +\alpha)}{\Gamma(|\mathcal{A}_{k'}| + \alpha)\Gamma(|\mathcal{A}_l|+\alpha)}\right) \\
  + &\sum_{g \leq K}\sum_{u \leq U}\log(L^*_{k'gu}) + \sum_{g \leq K}\sum_{u \leq U} \log(L^*_{lgu}) \\
  + &\sum_{k \leq K}\sum_{u \leq U}\log(L^*_{kk'u})+\sum_{k \leq K}\sum_{u \leq U}\log(L^*_{klu})\\
  - &\sum_u (\log(L^*_{k'k'u}) + \log(L^*_{k'lu}) +\log(L^*_{lk'u}) + \log(L^*_{llu})) \\
  - &\sum_{g \leq K}\sum_{u \leq U}\log(L_{k'gu}) - \sum_{g \leq K}\sum_{u \leq U} \log(L_{lgu}) \\
  - &\sum_{k \leq K}\sum_{u \leq U}\log(L_{kk'u})- \sum_{k \leq K}\sum_{u \leq U}\log(L_{klu})\\
  + &\sum_u (\log(L_{k'k'u}) + \log(L_{k'lu}) +\log(L_{lk'u}) + \log(L_{llu})),
  \end{split}
 \end{align}
\normalsize 
where $L_{kgu}$ is the term inside the product on the right hand side of equation \eqref{eq:T1} and $\mathbf{z^*}$ and $L_{kdu}^*$ refer to new configuration where the node $i$ in in $\mathcal{A}_l$.

 \normalsize
 When merging clusters $\mathcal{A}_{k'}$ and $\mathcal{A}_l$ into the cluster $\mathcal{A}_l$, the change in the ICL can be expressed as follows:
 \small
 \begin{align}
 \label{eq:merge}
 \begin{split}
 \Delta_{k'\rightarrow l} :=& ICL(\mathbf{z}^*, K-1)- ICL(\mathbf{z},K) =\\
 =&\log\left(\frac{p(\mathbf{z^*}| K-1)}{p(\mathbf{z}| K)} \right)+ \\
 + & \sum_{g \leq K}\sum_{u \leq U}(\log(L^*_{lgu}) + \log(L^*_{klu})) - \sum_u \log(L^*_{llu})\\
  -&\sum_{g \leq K}\sum_{u \leq U}\log(L_{k'gu}) - \sum_{g \leq K}\sum_{u \leq U} \log(L_{lgu}) \\
 - &\sum_{k \leq K}\sum_{u \leq U}\log(L_{kk'u})- \sum_{k \leq K}\sum_{u \leq U}\log(L_{klu})\\
  +&\sum_u (\log(L_{k'k'u}) + \log(L_{k'lu}) +\log(L_{lk'u}) + \log(L_{llu})).
 \end{split} 
 \end{align}
\normalsize	

When working with model \textbf{B}, we need to initialize $D_{max}$ and $\mathbf{y}$. Then a shuffled sequence of time intervals $I_1, \dots, I_U$ is considered and each interval is swapped to the time cluster leading to the highest increase in the ICL (\textbf{GE} for time intervals). When no further increase in the ICL is possible, we look for possible merges between time clusters in the attempt to increase the ICL (\textbf{GM} for time intervals). Formulas to directly assess the increase in the ICL can be obtained, similar to those for nodes swaps and merges.
In case of model \textbf{B}, different strategies are possible to optimize the ICL:
\begin{enumerate}
\item \textbf{GE} + \textbf{GM} for nodes at first and then for times (we will call this strategy \textbf{TN}, henceforth).
\item \textbf{GE} + \textbf{GM} for time intervals at first and then for nodes (\textbf{NT} strategy).
\item An hybrid strategy, involving alternate switching of nodes and time intervals (\textbf{M} strategy).      

\end{enumerate}
 We will provide details about the chosen strategy case by case in the following.       

\section{Experiments}\label{sec:Exp}
In  this section,  experiments on  both  synthetic and  real data  are
provided. All running times are measured on a twelve cores Intel Xeon server with 92 GB of
main memory running a GNU Linux operating system, the greedy algorithm
described  in  Section  \ref{par:GS}   being  implemented  in  C++.  A
Euclidean hierarchical clustering algorithm was used to initialize the
labels and $K_{max}$ was set to $N/2$. 

In the following, we call  TSBM the temporal SBM  we propose and we refer to the optimization algorithm described in the previous section as greedy ICL.

\subsection{Simulated Data}
\subsubsection{First Scenario} 
\begin{figure*}[ht]
\centering
\begin{subfigure}{.5\textwidth}
  \centering
  \includegraphics[width=\linewidth]{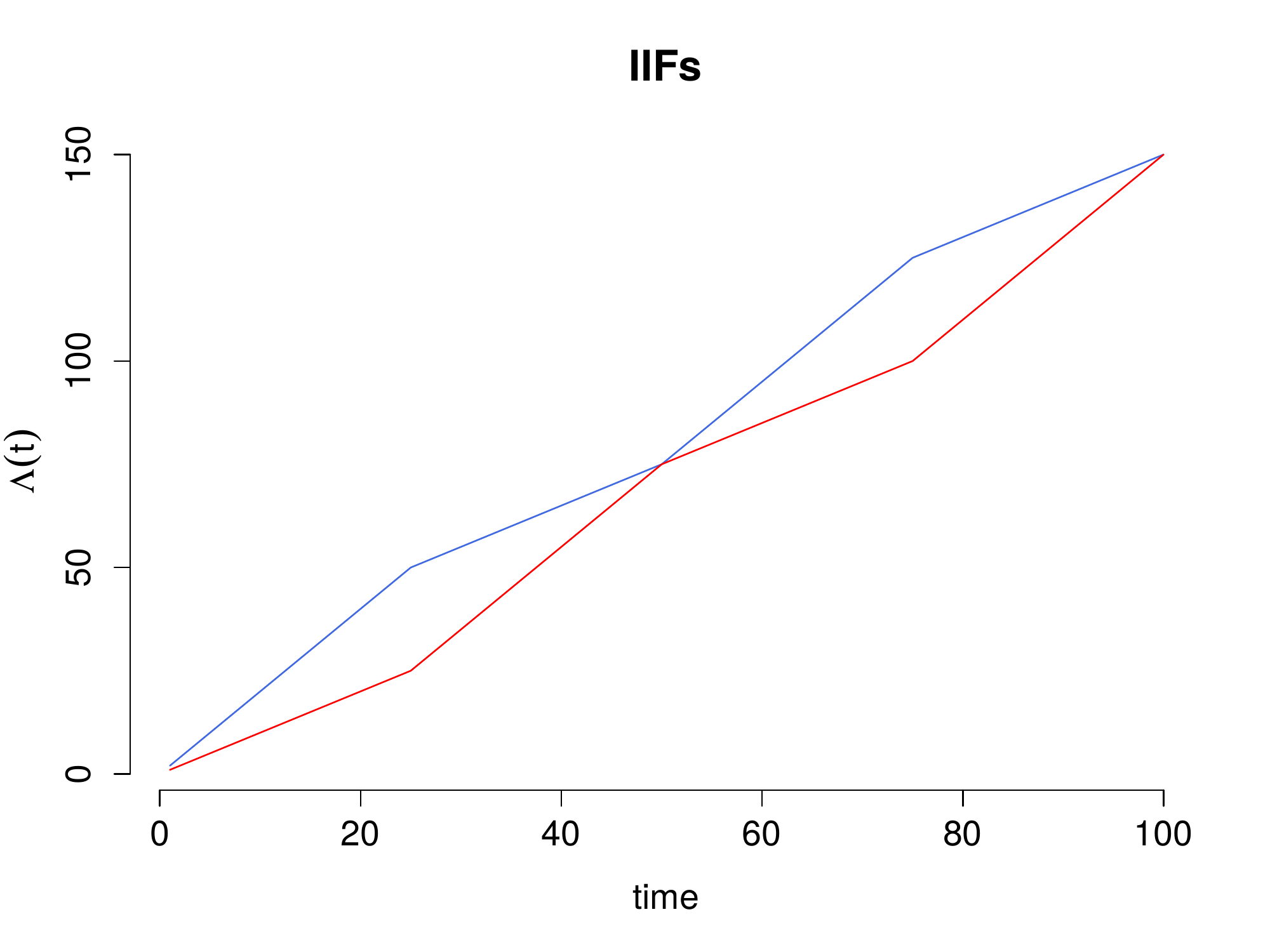}
  \captionsetup{format=hang}
  \subcaption{ }
  \label{fig:IIFs}
\end{subfigure}%
\begin{subfigure}{.5\textwidth}
  \centering
  \includegraphics[width=\linewidth]{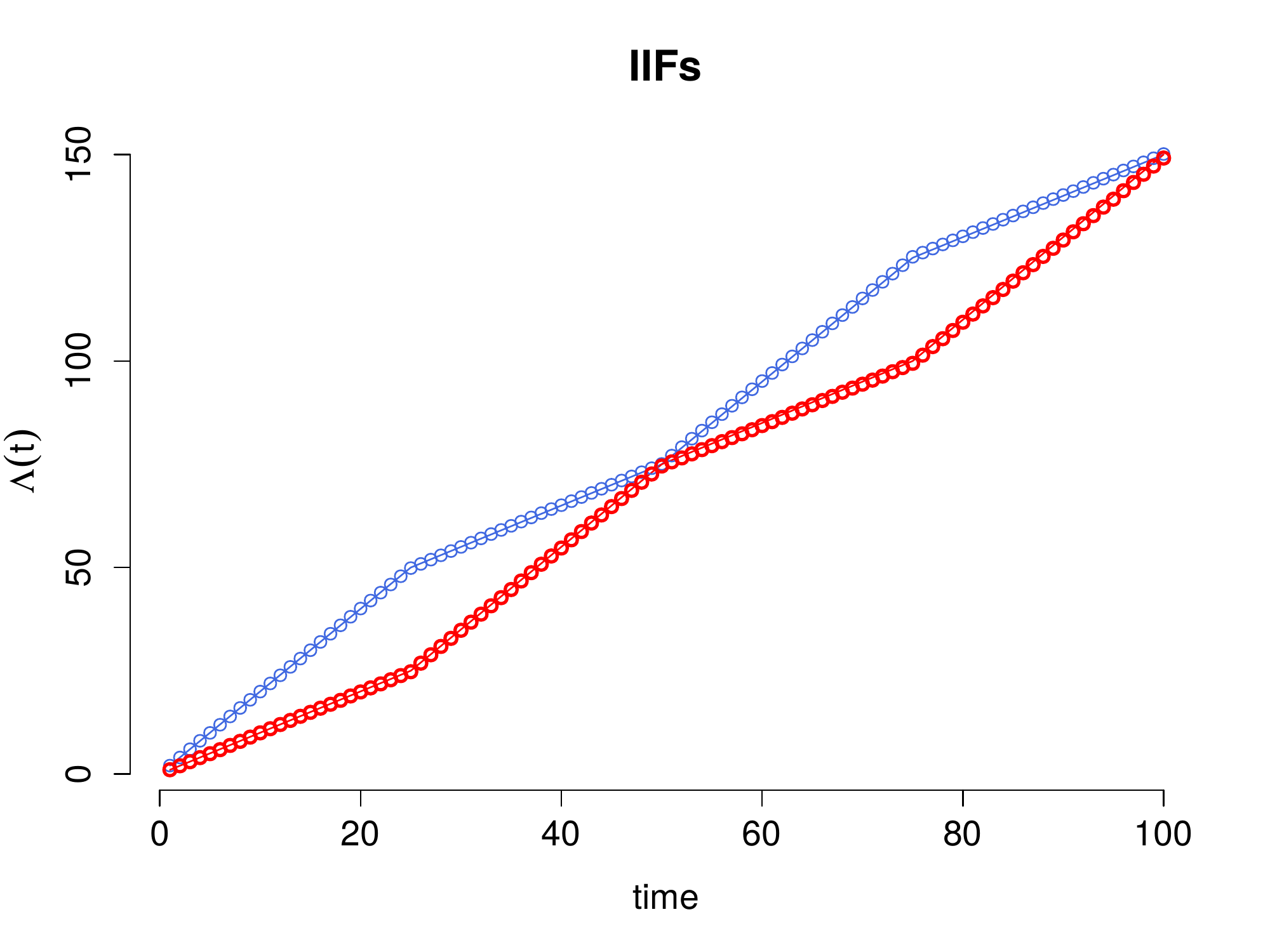}
  \captionsetup{format=hang}
  \subcaption{}
  \label{fig:IIFs_estim}
\end{subfigure}
\caption{Real \ref{fig:IIFs} and estimated \ref{fig:IIFs_estim} integrated intensity functions (IIFs) according to the considered generative model ($\psi=2$). In blue we have $\Lambda_1(t)$, for $\psi=4$, in red $\Lambda_2(t)$.}
\end{figure*}

We  start by  investigating  how the  proposed approach  can  be used  to
efficiently estimate  the vector $\mathbf{z}$ of  labels in situations
where the standard  SBM fails. Thus, we  simulate interactions between
50 $(N)$ nodes,  grouped in two hidden clusters $\mathcal{A}_1$  and $\mathcal{A}_2$, over
100  $(U)$ time  intervals  of unitary  length.  The generative  model
considered for the simulations depends  on two time clusters $\mathcal{C}_1$ and
$\mathcal{C}_2$  containing  a  certain  number  of  time  intervals  $I_1,\dots
I_U$. If $I_u$ is in $\mathcal{C}_1$ then $Y_{ij}^{I_u}$ is drawn from a Poisson
distribution $\mathcal{P}(P_{z_i z_j})$. Otherwise,  $Y_{ij}^{I_u}$ is drawn from a Poisson
distribution $\mathcal{P}(Q_{z_i z_j})$. The  matrices $P$ and $Q$ are
given by
\begin{equation*}
     P=
  \begin{pmatrix}
     \psi &  1  \\
     1 &  \psi  \\
  \end{pmatrix}
\qquad\text{and}\qquad
     Q=
  \begin{pmatrix}
     1 &  \psi  \\
     \psi &  1  \\
  \end{pmatrix},
\end{equation*}
where $\psi$ is a free parameter in $[1, \infty)$. When this parameter is equal to 1, we are in a degenerate case and there is not any structure to detect: all the nodes are placed in the same, unique cluster. The higher $\psi$, the stronger the $\emph{contrast}$ between the interactions pattern inside and outside the cluster. 
In this paragraph, $\psi$  is set equal to 2 and the proportions of the clusters are set equal ($\boldsymbol{\omega}=(1/2, 1/2)$). The number of time intervals assigned to each time cluster is assumed to be equal to $U/2$. In the following, we consider 
\begin{align*}
\mathcal{C}_1:=&\{I_1, \dots, I_{25}\}\cup \{I_{51},\dots,I_{75}\}, \\
\mathcal{C}_2:=&\{I_{26}, \dots, I_{50}\}\cup \{I_{76},\dots,I_{100}\}.
\end{align*}
This generative model defines two integrated intensity functions (IIFs), say $\Lambda_1(t)$ and $\Lambda_2(t)$. The former is the IIF of the Poisson processes counting interactions between nodes sharing the same cluster, the latter is the IIF of the Poisson processes counting interactions between vertices in different clusters. These IIFs can be observed in Figure \ref{fig:IIFs}.

%

A tensor $Y$, with dimensions $N \times N \times U$, is drawn. Its $(i,j,u)$ component is the sampled number of interactions from node $i$ to node $j$ over the time interval $I_u$.
Moreover, sampled interactions are aggregated over the whole time horizon to obtain an adjacency matrix. In other words, each tensor is integrated over its third dimension.
We compared the greedy ICL  algorithm with the Gibbs sampling approach  introduced by \cite{nouedoui2013}.  The former was run on the tensor $Y$
(providing estimates in 11.86 \emph{seconds} on average) the latter on
the corresponding  adjacency matrix.  This experiment was  repeated 50
times and estimates of random vector $\mathbf{z}$ were provided at each  iteration. Each estimate $\hat{\mathbf{z}}$ is compared with the true $\mathbf{z}$ and an adjusted rand index \citep[ARI][]{rand1971objective} is computed. This index takes values between zero and one, where one corresponds to the perfect clustering (up to label switching).  
\begin{Remark}
the true  structure is
always  recovered by  the  TSBM: 50 unitary values of the ARI are obtained. Conversely,  the  standard SBM  never succeeds in recovering any hidden structures present in the data (50 null ARIs are obtained). This can  easily  be  explained  since  the  time  clusters  have  opposite
interaction  patterns, making  them hard  to uncover  when aggregating
over time. 
\end{Remark}

Relying on an efficient estimate  of $\mathbf{z}$, the
two integrated intensity functions can be estimated through the estimator in equation
\eqref{eq:linear_estimator}. Results can be observed in Figure \ref{fig:IIFs_estim}, where the estimated functions (coloured dots) overlap the real functions \ref{fig:IIFs}.

\paragraph{Over fitting}
We now illustrate  how the model discussed so far  fails in recovering
the true  vector $\mathbf{z}$ when  the number of time  intervals (and
hence  of  free parameters)  grows. 
We consider the same generative model of the previous paragraph, with a lower $\psi$:
\begin{equation*}
     P=
  \begin{pmatrix}
     1.4 &  1  \\
     1 &  1.4  \\
  \end{pmatrix}
\qquad\text{and}\qquad
     Q=
  \begin{pmatrix}
     1 &  1.4  \\
     1.4 &  1  \\
  \end{pmatrix}.
\end{equation*}
Despite the lower contrast (from $2$ to $1.4$ in $P$ and $Q$), with $U=100$ and time sub-intervals of
unitary length, the TSBM model still always recovers the true vector
$\mathbf{z}$.  Now  we consider  a  finer  partition of  $[0,100]$  by
setting $U=1000$ and $\Delta_u=0.1$ as well as scaling the intensity matrices as follows
\begin{equation*}
     \tilde{P}:=
  \begin{pmatrix}
     0.14 &  0.1  \\
     0.1 &  0.14  \\
  \end{pmatrix}
\qquad\text{and}\qquad
     \tilde{Q}=
  \begin{pmatrix}
     0.1 &  0.14  \\
     0.14 &  0.1  \\
  \end{pmatrix}.
\end{equation*}
Moreover, we set
\begin{equation*} 
 \mathcal{C}_1:=\{I_1, \dots, I_{250}\} \cup \{I_{501}, \dots, I_{750}\} 
\end{equation*}
and $\mathcal{C}_2$ is the complement of $\mathcal{C}_1$, as previously. Finally, we sampled 50 dynamic graphs over the interval $[0,100]$ from
the corresponding generative model. Thus, each graph is characterized by a sampled tensor $Y$. 

Unfortunately, the  model is not  robust to such changes.  Indeed, when
running  the greedy  ICL algorithm  on  each sampled  tensor $Y$,  the
algorithm  does not  see any  community  structure and  all nodes  are
placed  in the  same  cluster. This  leads  to a  null  ARI, for  each
estimation. As mentioned in  paragraph \ref{par:GS}, the ICL penalizes
the number of  parameters and since the tensor $\pi$  has dimension $K
\times  K \times  U$, for  a fixed  $K$, when  moving from  the larger
decomposition ($U=100$)  to the  finer one  ($U=1000$), the  number of
free parameters in the model is approximatively\footnote{The dimension
  of the vector $\boldsymbol{\omega}$  does not change.} multiplied by
10. The increase we observe in the likelihood, when
increasing the number of clusters of nodes from $K=1$ to $K=2$, is not
sufficient  to  compensate the  penalty  due  to  the high  number  of
parameters and hence the ICL decreases. Therefore, the maximum is taken for $K=1$ and a single cluster is detected.

Model  \textbf{B}  allows to  tackle  this  issue. When  allowing  the
integrated  intensity functions  $\Lambda_1(t)$ and  $\Lambda_2(t)$ to
grow at the same rate on each interval $I_u$ belonging to the same time cluster $\mathcal{C}_d$, we basically reduce the third dimension of the tensor $\pi$ from $U$ to $D$.

\begin{figure}[ht]
 \centering
 \includegraphics[width=.9\linewidth]{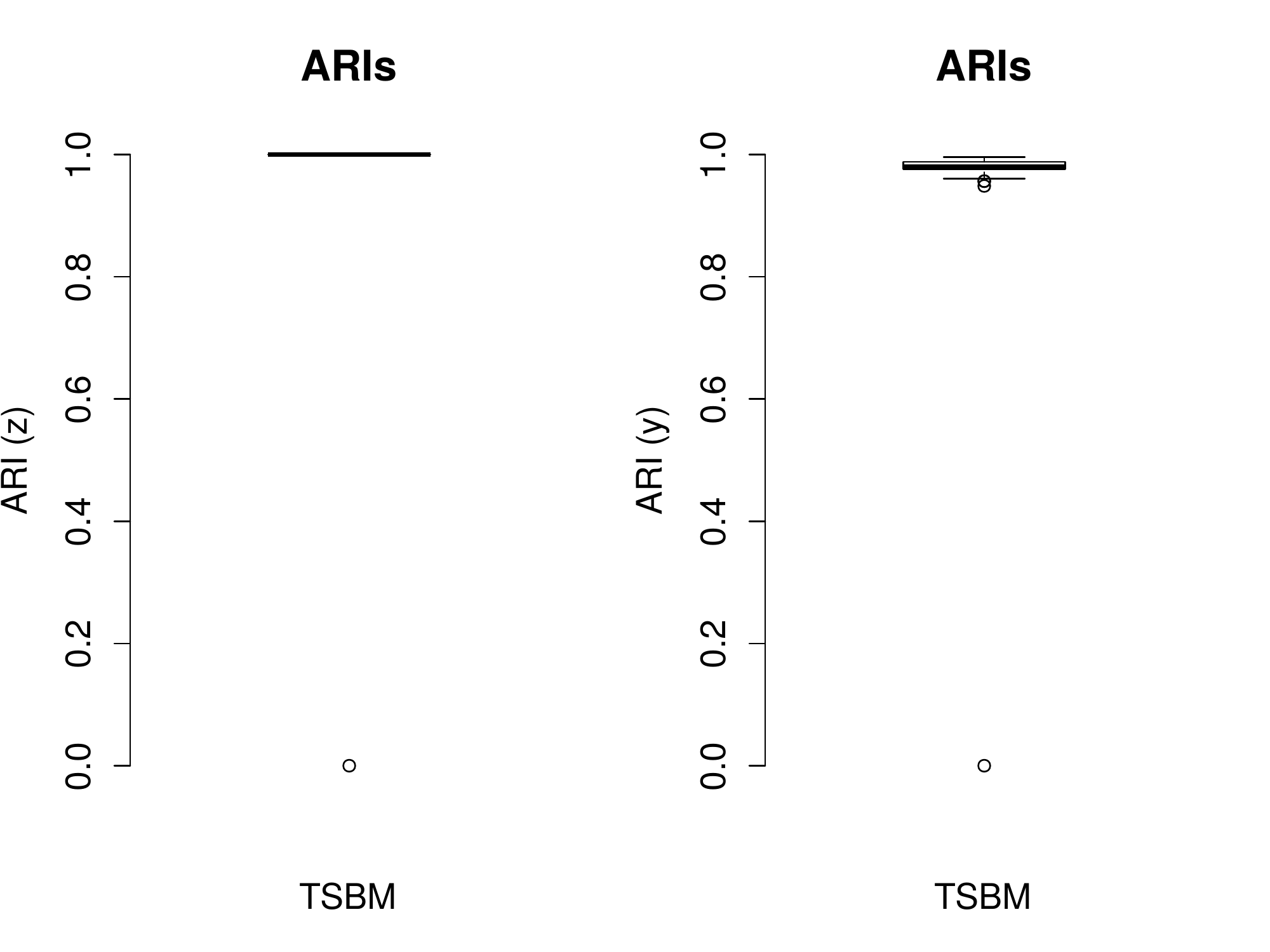}
 \caption{Box plots for both clusterings of nodes and time intervals: 50 dynamic graphs were sampled according to the considered generative model, estimates of $\mathbf{z}$ and $\mathbf{y}$ are provided by the greedy ICL (model B). }
\label{fig:MB1}
\end{figure}

The greedy ICL algorithm for Model  \textbf{B} was run on each sampled
tensor $Y$,  providing estimates  of $\mathbf{z}$ and  $\mathbf{y}$ in
$2.38$ \emph{minutes}, on average.
A hierarchical  clustering algorithm was  used to initialize  the time
labels $\mathbf{y}$, and  the initial number of time  clusters was set
to $D_{max}=\sqrt{U}$.  In an  attempt to  avoid convergence  to local
maxima,  ten estimates  are built  for  each tensor  and the  estimate
leading to the  best ICL is finally retained. The  adjusted rand index
is used to evaluate the clustering, as previously, and the results are
presented as box plots in Figure \ref{fig:MB1}.
Note that the results were obtained through the optimization strategy \textbf{TN}. The other two strategies described in section \ref{par:GS}, namely the \textbf{NT} strategy and the \textbf{M} strategy, led to similar results in terms of final ICL and ARIs.

\subsubsection{Second Scenario} 

Since the node clusters are fixed over time, the TSBM model can be seen as an alternative to a standard SBM to estimate the label vector $\mathbf{z}$. The previous scenario shows that the TSBM can recover the true vector $\mathbf{z}$ in situations where the SBM fails. In this paragraph we show how the TSBM and the SBM can sometimes have similar performances. 

\begin{figure*}[ht]
\centering
\begin{subfigure}{.5\textwidth}
  \centering
  \includegraphics[width=\linewidth]{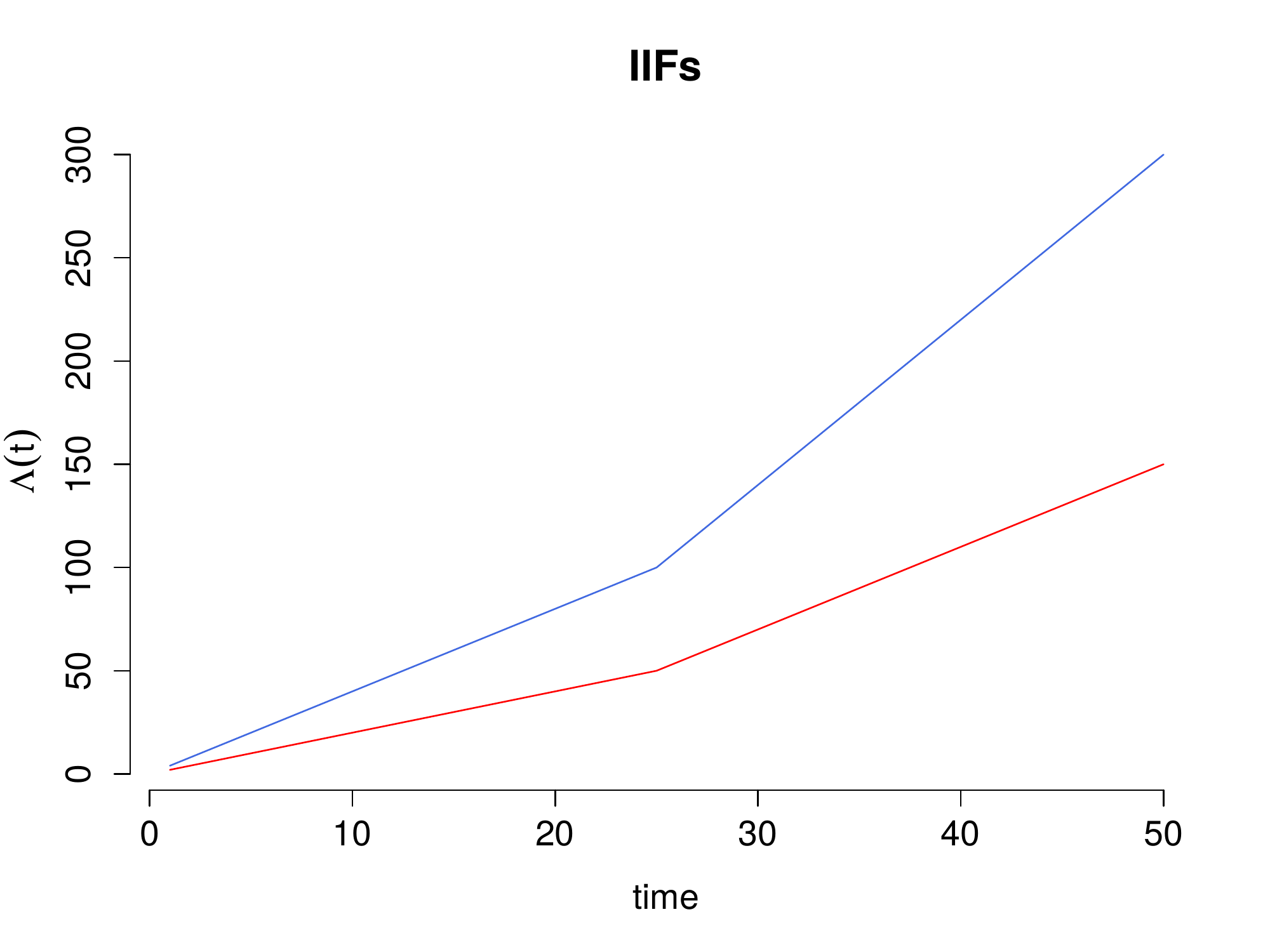}
  \captionsetup{format=hang}
  \subcaption{ }
  \label{fig:IIFs_2}
\end{subfigure}%
\begin{subfigure}{.5\textwidth}
  \centering
  \includegraphics[width=\linewidth]{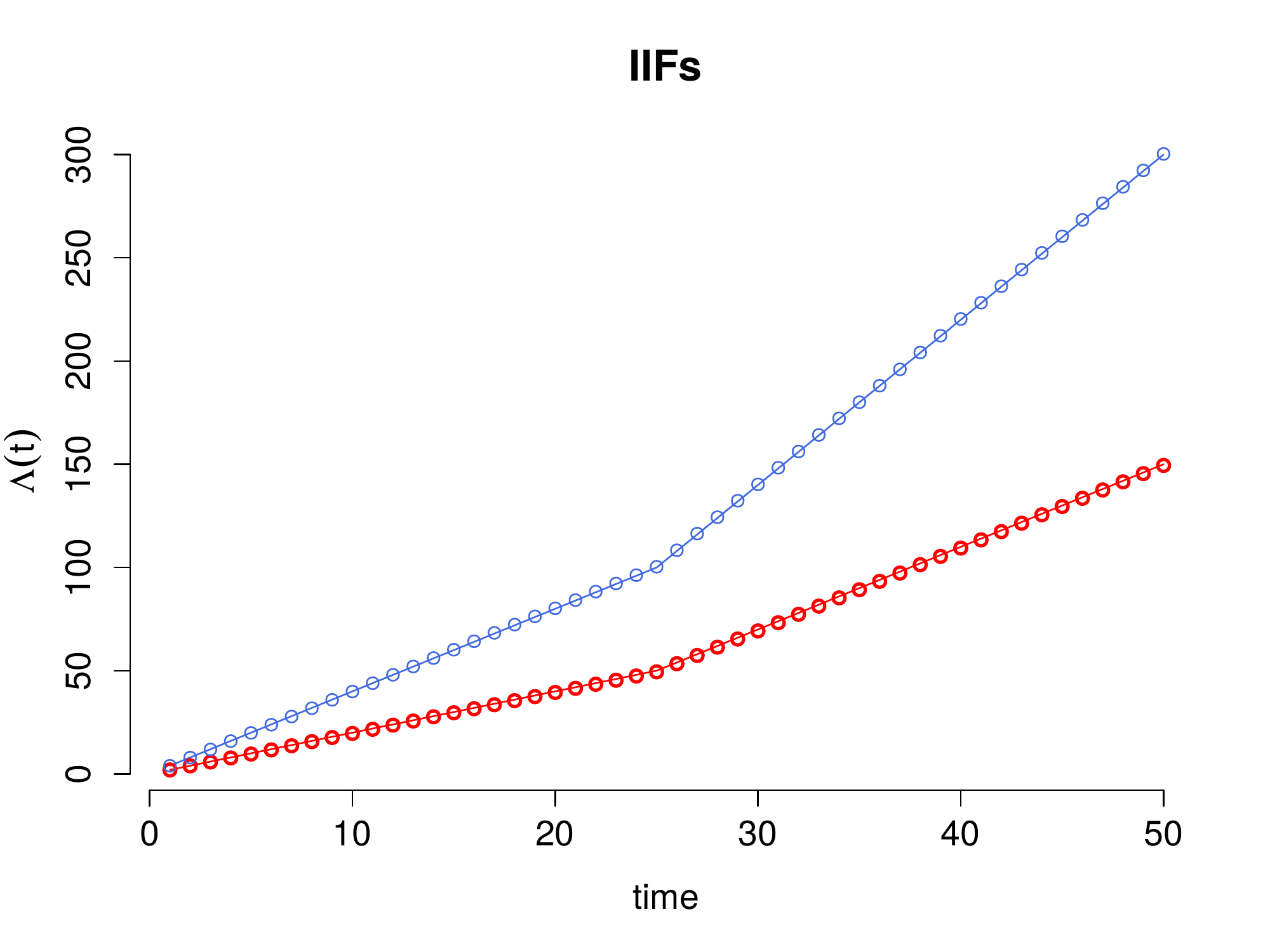}
  \captionsetup{format=hang}
  \subcaption{}
  \label{fig:IIFs_2_estim}
\end{subfigure}
\caption{Real \ref{fig:IIFs_2} and estimated \ref{fig:IIFs_2_estim} integrated intensity functions (IIFs) according to the considered generative model. In blue we have $\Lambda_1(t)$, for $\psi=4$, in red $\Lambda_2(t)$.}
\end{figure*}

We considered dynamic graphs with 50 $(N)$ nodes and 50 $(U)$ time intervals 
\begin{equation*}
 I_1, \dots, I_{50}.
\end{equation*}
These time intervals are grouped in two time clusters $\mathcal{C}_1$ and $\mathcal{C}_2$, the former containing the first 25 time intervals, the latter the last 25 time intervals. If $I_u$ is in $\mathcal{C}_1$ then $Y_{ij}^{I_u}$ is drawn from a Poisson
distribution $\mathcal{P}(P_{z_i z_j})$. Otherwise,  $Y_{ij}^{I_u}$ is drawn from a Poisson
distribution $\mathcal{P}(2P_{z_i z_j})$.
The $P$ matrix is given by
\begin{equation*}
     P=
  \begin{pmatrix}
     \psi &  2 \\
      2 & \psi 
  \end{pmatrix}
\end{equation*}
and $\psi$ is a free parameter in $[2, +\infty)$. 
Hence,  we  have two  different  integrated  intensity functions,  say
$\Lambda_1(t)$  and  $\Lambda_2(t)$ with  the  same  roles as  in  the
previous section. These two functions are plotted in Figure \ref{fig:IIFs_2}, for a value of $\psi=4$.  

\begin{figure*}[ht]
\centering
\begin{subfigure}{.5\textwidth}
  \centering
  \includegraphics[width=\linewidth]{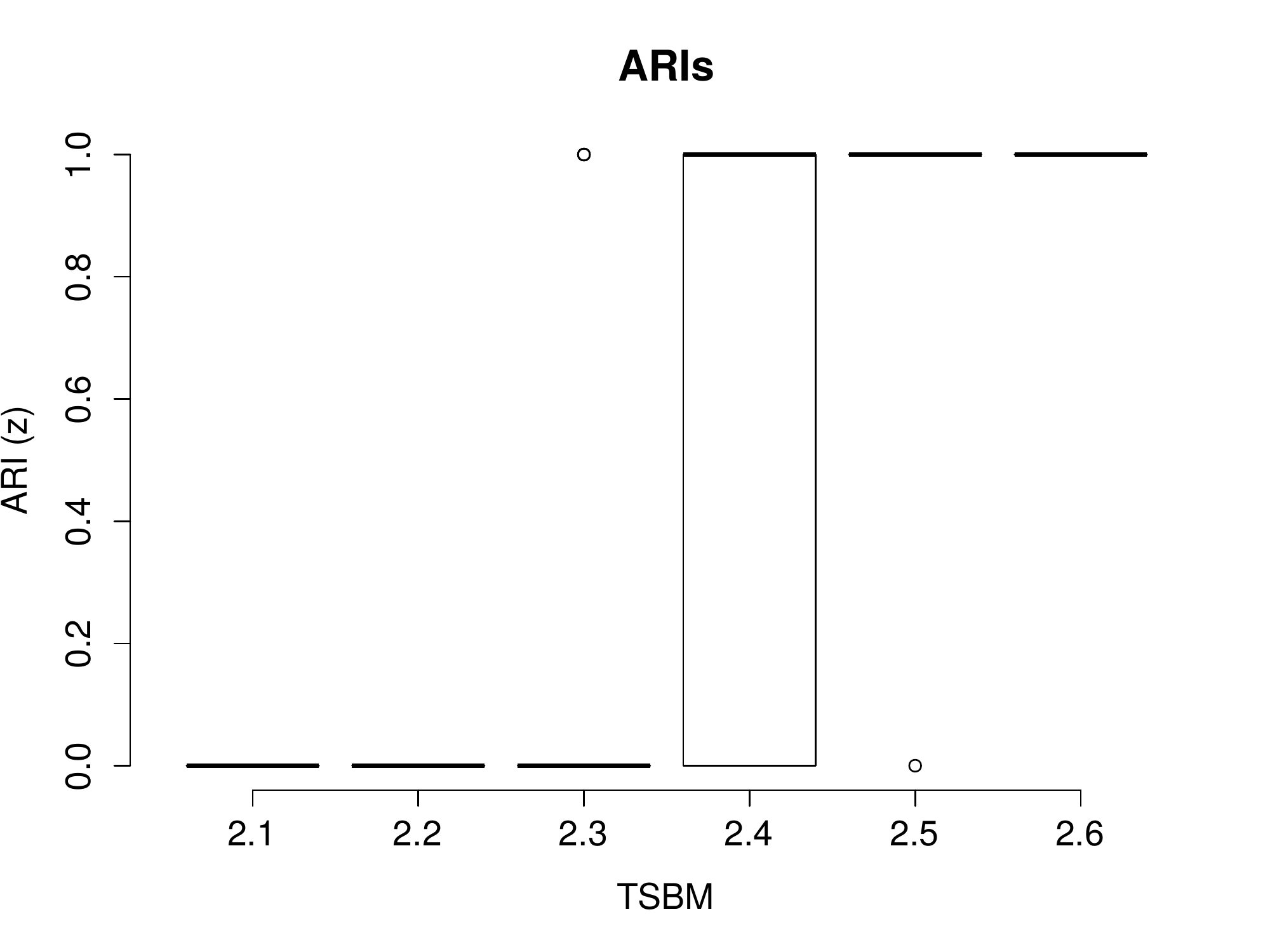}
  \captionsetup{format=hang}
  \subcaption{\footnotesize ARIs obtained by greedy ICL.}
  \label{fig:Ari_TSBM}
\end{subfigure}%
\begin{subfigure}{.5\textwidth}
  \centering
  \includegraphics[width=\linewidth]{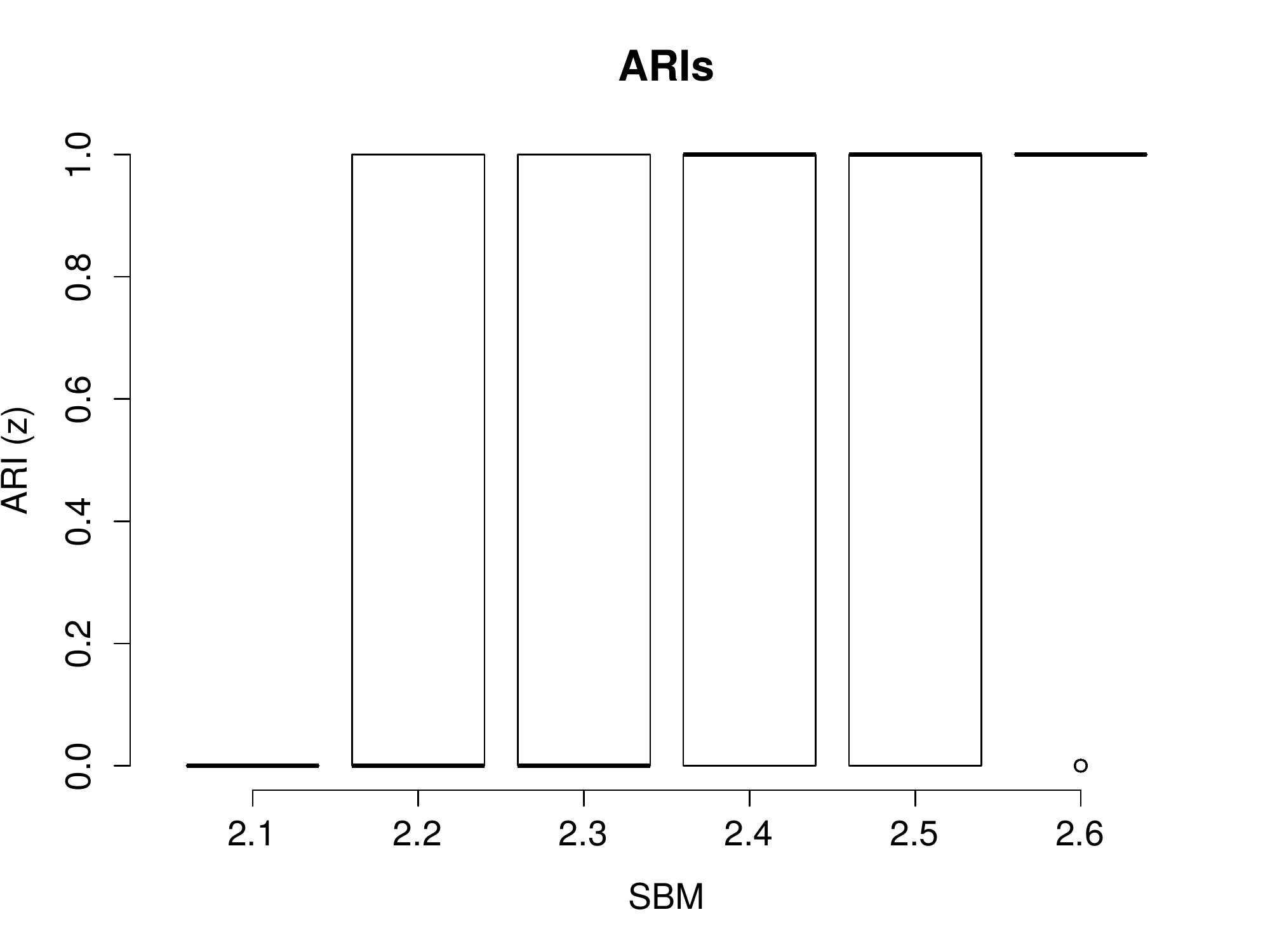}
  \captionsetup{format=hang}
  \subcaption{\footnotesize  ARIs  obtained  with the  Gibbs  sampling
    procedure for SBM.}
  \label{fig:Ari_SBM}
\end{subfigure}
\caption{Box plots of ARIs for different levels of contrast ($\psi$). We compare the proposed model with a standard SBM.}
\label{fig:ari_s}
\end{figure*}

We investigated six values for the parameter $\psi$ 
\begin{equation*}
 \{2.1, 2.2, 2.3, 2.4,2.5,2.6\}.
\end{equation*}
 For each  value of $\psi$,  we sampled  50 tensors $Y$,  of dimension
 $(50  \times  50  \times  50)$, according  to  the  generative  model
 considered.  Interactions  are  aggregated  over  the  time  interval
 $[0,50]$ to obtain adjacency matrices. We ran the greedy ICL algorithm
 on  each  tensor and  the  Gibbs  sampling  (SBM) algorithm  on  each
 adjacency matrix. For  the greedy ICL algorithm,  estimates of vector
 $\mathbf{z}$  were   obtained  in  a   mean  running  time   of  5.52
 \emph{seconds}. As previously, to  avoid convergence to local maxima,
 ten different estimates are built for each tensor, the one leading to
 the highest ICL being retained. The results are presented as box plots in Figure \ref{fig:ari_s}.
Although the SBM leads to slightly better clustering results for small values of $\psi$ (2.2, 2.3) and the TSBM for higher values of $\psi$ (2.5, 2.6), we observe that the two models have quite similar performances (in terms of accuracy) in this scenario.  

\begin{figure*}[ht]
 \centering
 \includegraphics[width=.9\linewidth]{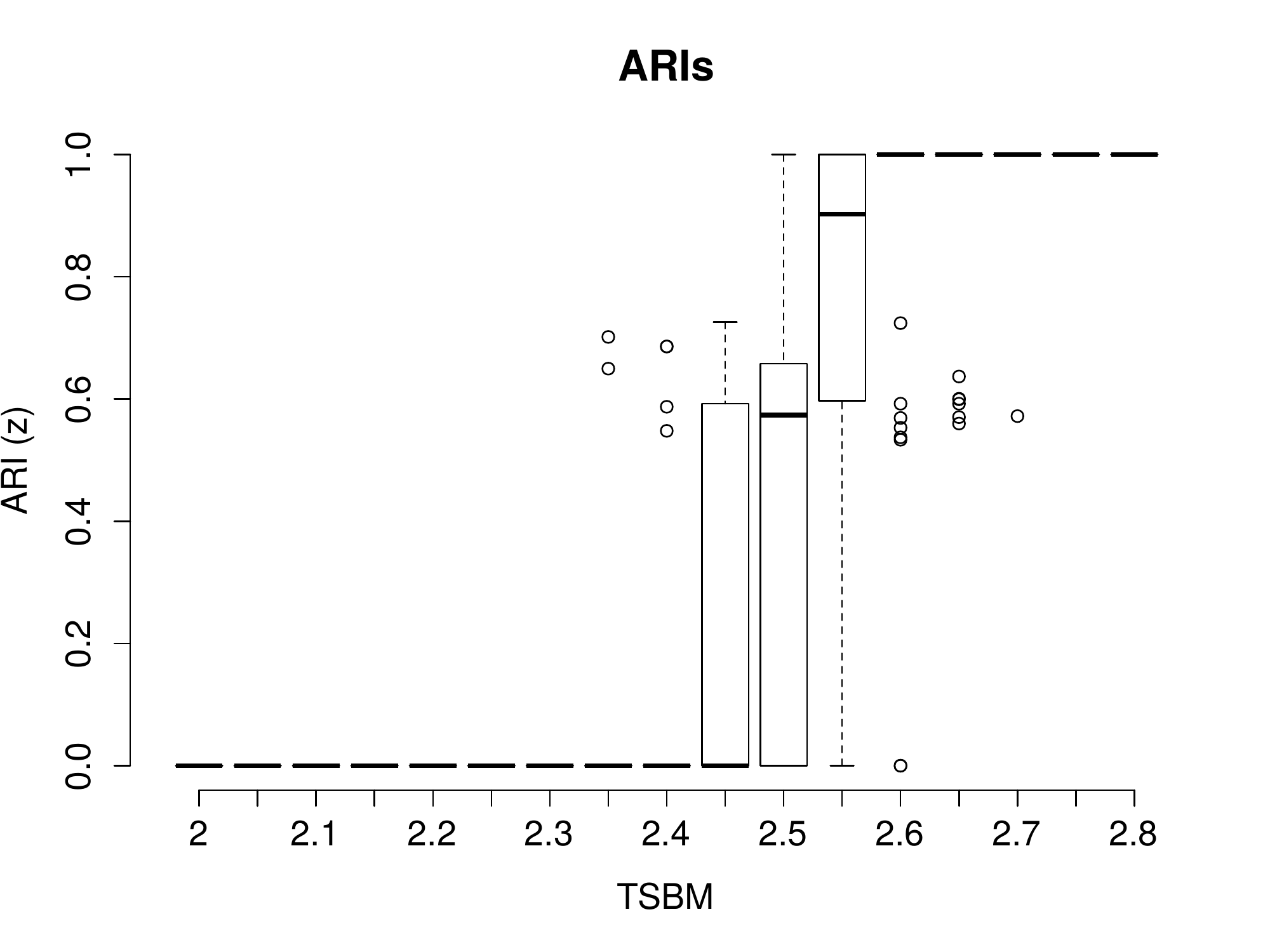}
 \caption{Box plots of ARIs for different levels of contrast ($\psi$). Data have been sampled by non-homogeneous Poisson processes counting interactions in a dynamic graph whose nodes are grouped in three clusters and interactivity patterns vary across two time clusters. }
\label{fig:repeat}
\end{figure*}

To provide some intuitions about the scalability (see next paragraph) of the proposed approach we repeated the previous experiment by setting $K=3$ clusters, corresponding to the following connectivity matrix:
\begin{equation*}
     P=
  \begin{pmatrix}
     \psi &  2   & 2\\
      2   & \psi & 2\\
      2   & 2    & \psi
  \end{pmatrix}.
\end{equation*}
The assignment of the time intervals to the time clusters is unchanged as well as the connectivity pattern on each time cluster are unchanged.
The contrast parameter $\psi$ takes values in the set $\{2, 2.5,2.10, \dots, 2.8 \}$ and 50 dynamic graphs were sampled, according to 
the described settings, for each value of $\psi$. We ran the TSBM on each dynamic graph obtaining 50 estimates of the labels vector $\mathbf{z}$ (one for each $\psi$) and box and whiskers plots for each group of ARIs can be seen in Figure \ref{fig:repeat}.  
By comparing this figure with Figure \ref{fig:Ari_TSBM}, we can see that the model needs a slight higher contrast to fully recover the true structure. Actually, when increasing the number of clusters without increasing the number of nodes, the size of each cluster decreases (on average) and since the estimator of $\mathbf{z}$ we are using is related to the ML estimator, we can imagine a slower convergence to the true value of $\mathbf{z}$.

\subsubsection{Scalability}
A full scalability analysis of the proposed algorithm as well as the convergence properties of the proposed estimators are outside the scope of this paper. Nonetheless, in appendix we provide details about the computational complexity of the greedy-ICL algorithm. Future works could certainly be devoted to improve both the algorithm efficiency and scalability through the use of more sophisticated data structures.

\subsection{Real data}

The dataset used in this section was collected during the \textbf{ACM Hypertext} conference held in Turin, June 29th - July 1st 2009. 
We focus on the first conference day (24 hours) and consider a dynamic network with 113 $(N)$ nodes (conference attendees) and 96 $(U)$ time intervals (the consecutive quarter-hours in the period: 8am of June 29th  - 7.59am of June 30th). The network edges are the proximity face to face interactions between the conference attendees. An interaction is monitored when two attendees are face to face, nearer than 1.5 meters for a time period of at least 20 seconds\footnote{More informations about the way the data were collected can be found in \cite{Isella:2011qo} or visiting the website
\url{http://www.sociopatterns.org/data sets/hypertext-2009-dynamic-contact-network/ }.
}.
The data set we considered consists of several lines similar to the following one

\begin{center}
 \begin{tabular}{c|c|c|c}
  \hline
  \footnotesize\emph{ID$1$} & \footnotesize\emph{ID$2$} & \footnotesize\emph{Time Interval ($15m$)} & \footnotesize\emph{Number of interactions} \\
  \hline
   52 & 26 & 5 & 16 \\
  \hline 
 \end{tabular}
\end{center}
\medskip
It means that conference attendees 52 and 26, between 9am and 9.15am, have spoken for $16 \times 20s \approx 5m30s$.  

\begin{figure}[ht]
 \centering
 \includegraphics[width=.9\linewidth]{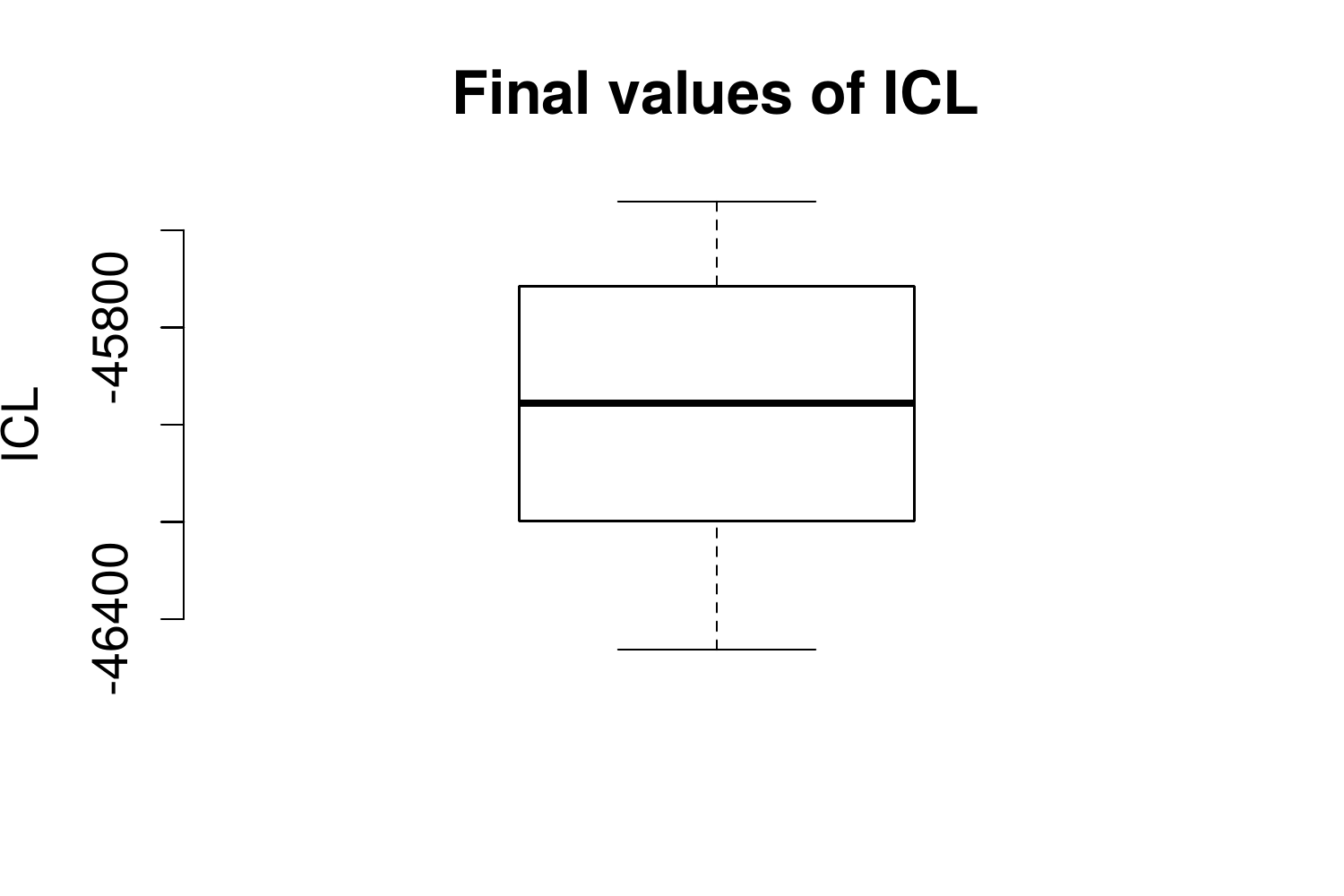}
 \caption{Box plot of the ten final values of the ICL produced by the greedy ICL algorithm for different initializations.}
\label{fig:final_icl_boxplot}
\end{figure}

We set $K_{max}=20$ and the vector $\mathbf{z}$ was initialized randomly: each node was assigned to a cluster following a multinomial distribution. The greedy algorithm was run ten times on the considered dataset, each time with a different initialization and estimates of $\mathbf{z}$ and $K$ were provided in 13.81 \emph{seconds}, on average. The final values of the ICL can be observed as box plots in Figure \ref{fig:final_icl_boxplot}	.

\begin{figure*}[ht]
\centering
\begin{subfigure}{.5\textwidth}
  \centering
  \includegraphics[width=\linewidth]{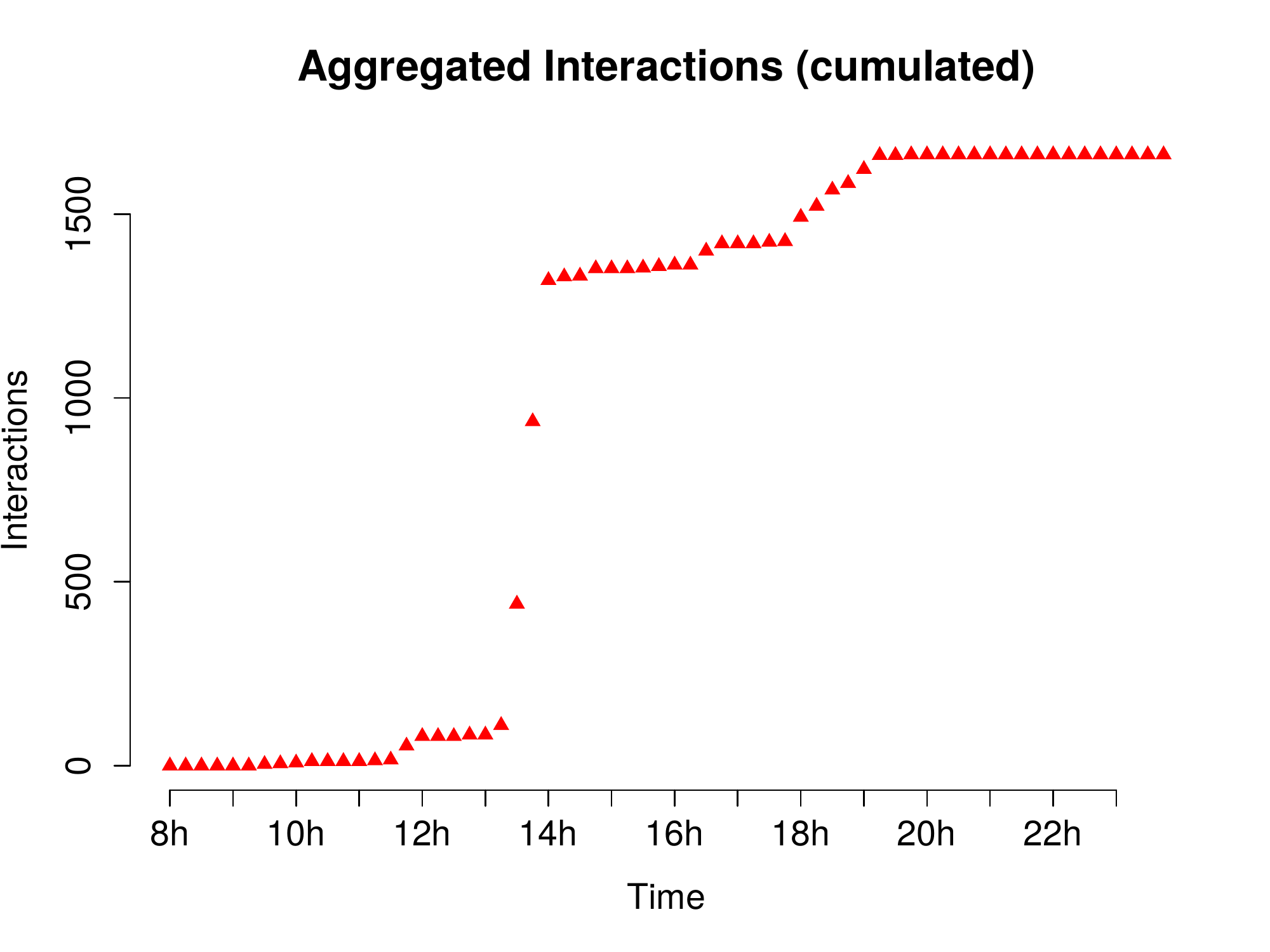} 
  \caption{\footnotesize Cumulated aggregated connections inside cluster $\mathcal{A}_4$.}
  \label{fig:Agginter4}
\end{subfigure}%
\begin{subfigure}{.5\textwidth}
  \centering
  \includegraphics[width=\linewidth]{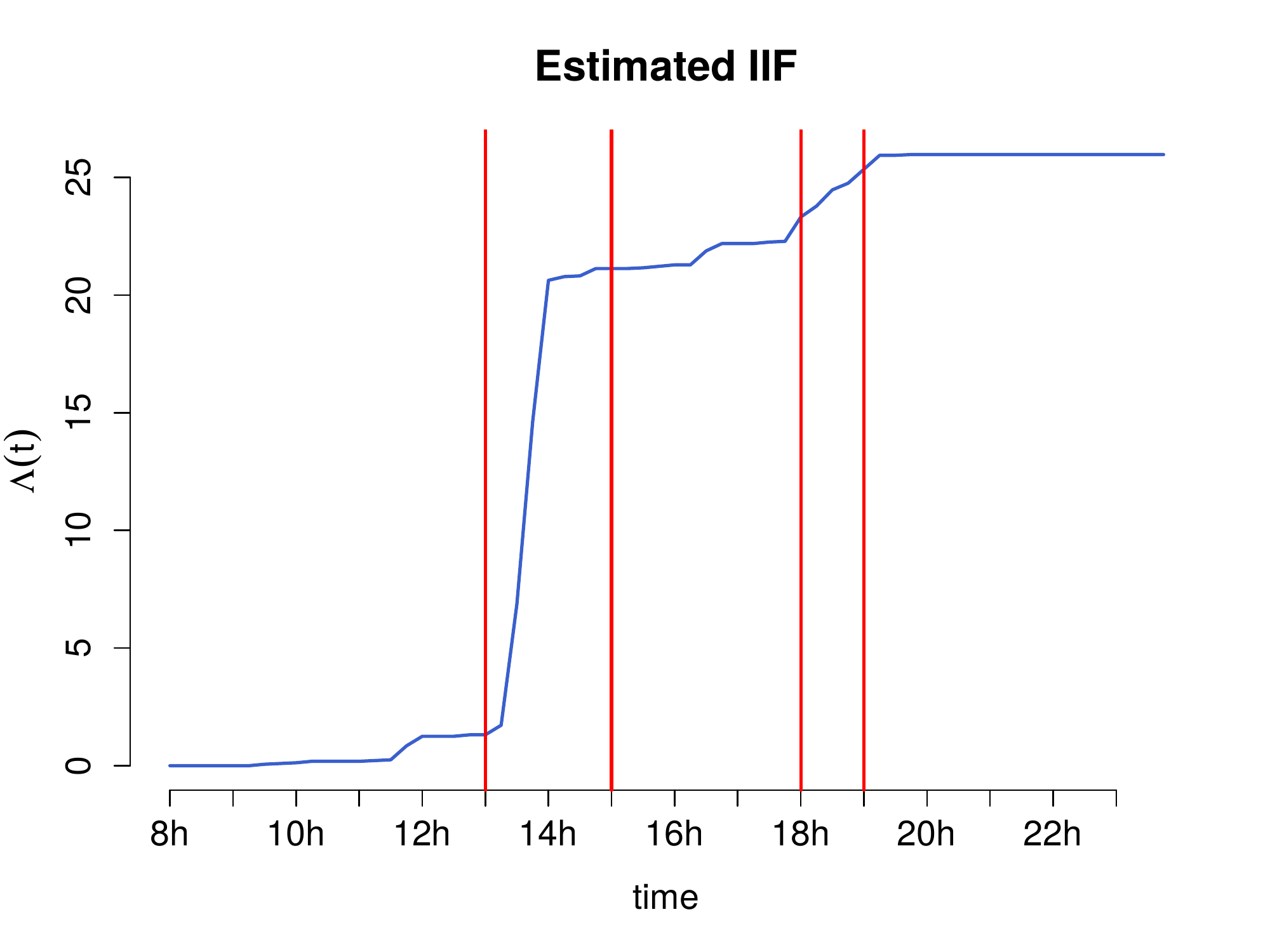}
  \caption{\footnotesize Estimated IIF for interactions inside cluster $\mathcal{A}_4$. }
  \label{fig:E_iif}
\end{subfigure}
\caption{\small in Figure \ref{fig:Agginter4}, cumulated aggregated connections for each time interval for cluster $\mathcal{A}_4$ . In Figure \ref{fig:E_iif} the estimated IIF for interactions inside cluster $\mathcal{A}_4$. Vertical red lines delimit the lunch break and the wine and cheese reception.}
\label{fig:MainFIgure}
\end{figure*}

 The estimates associated to the highest ICL correspond to 5 node clusters. In Figure \ref{fig:MainFIgure}, we focus on the cluster $\mathcal{A}_4$, containing 48 nodes. In Figure \ref{fig:Agginter4} we plotted the time cumulated interactions inside the cluster. As it can be seen the connectivity pattern for this cluster is very representative of the entire graph: between 13pm and 14pm and 18pm and 19.30pm there are significant increases in the interactions intensity.     
The estimated integrated intensity function (IIF) for interactions inside this cluster can be observed in Figure \ref{fig:E_iif}. The function has a higher slope on those time intervals where attendees in the cluster are more likely to have interactions. The vertical red lines delimit two important times of social gathering\footnote{More informations at \url{http://www.ht2009.org/program.php}.}: 
\begin{itemize}
 \item 13.00-15.00 - lunch break.
 \item 18.00-19.00 - wine and cheese reception.
\end{itemize}

\begin{figure*}[ht]
\centering
\begin{subfigure}{.5\textwidth}
  \centering
  \includegraphics[width=\linewidth]{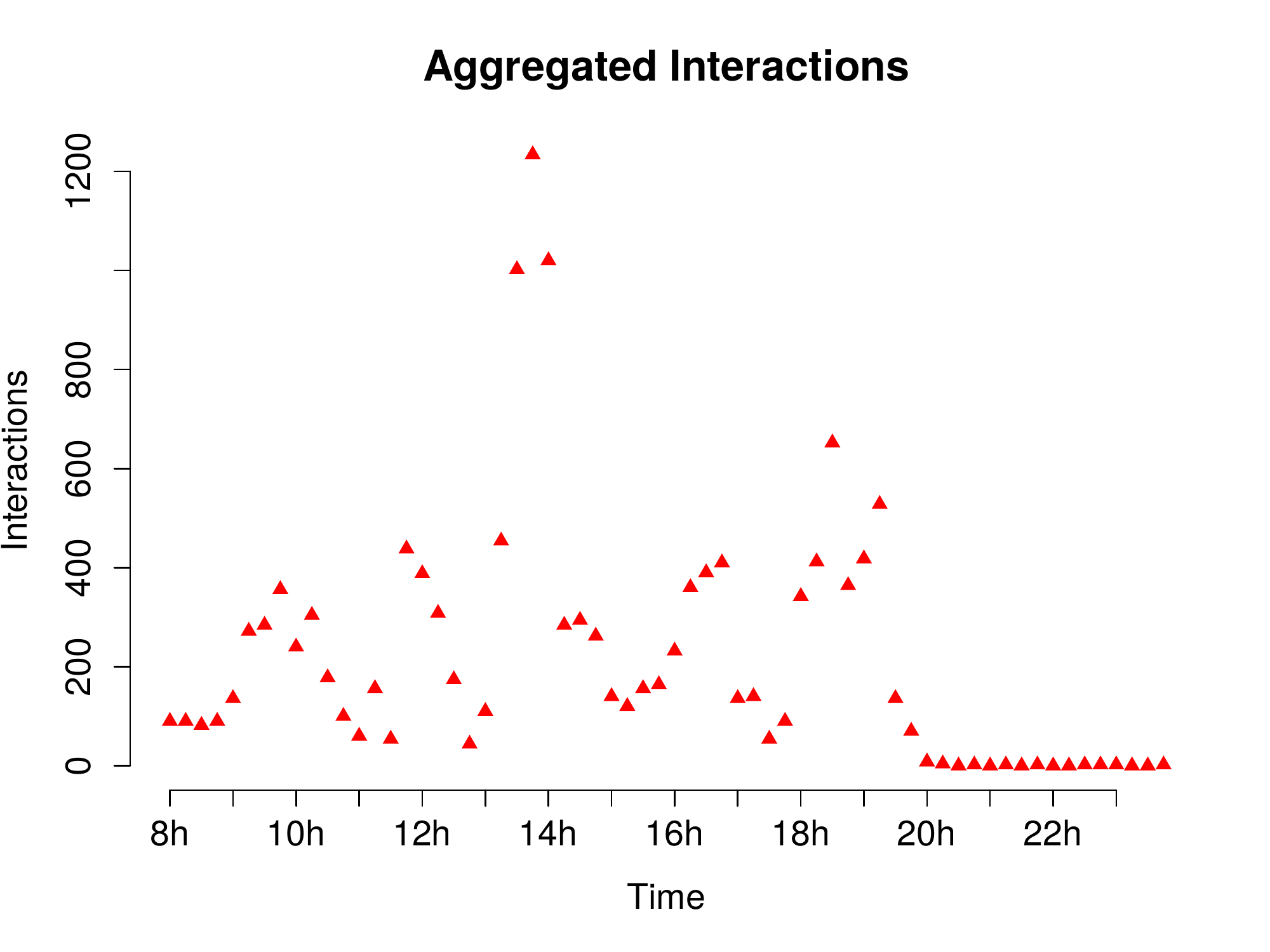} 
  \caption{\footnotesize Aggregated connections.}
  \label{fig:Agginter}
\end{subfigure}%
\begin{subfigure}{.5\textwidth}
  \centering
  \includegraphics[width=\linewidth]{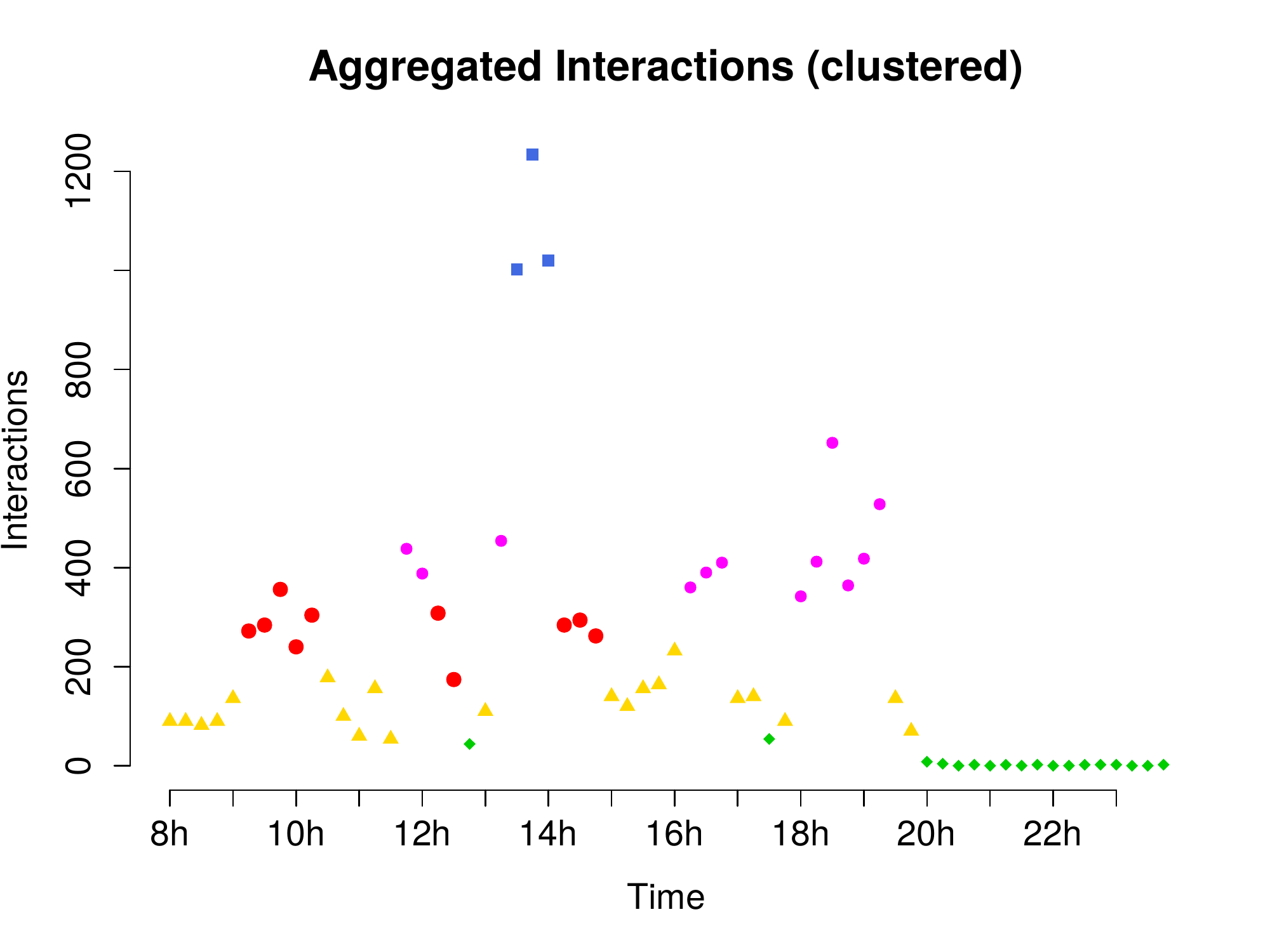}
  \caption{\footnotesize Clustered time intervals. }
  \label{fig:Agginter_clust}
\end{subfigure}
\caption{\small in Figure \ref{fig:Agginter}, aggregated connections for each time interval for the whole network. In Figure \ref{fig:Agginter_clust} interactions of the same form/color take  place on time intervals assigned to the same cluster (model \textbf{B}). }
\label{fig:MainFIgure2}
\end{figure*}

We conclude this section by illustrating how Model \textbf{B} can be used to assign time intervals on which interactions have similar intensity to the same time cluster. We run the greedy ICL algorithm for Model \textbf{B} on the dataset by using the optimization strategy \textbf{M} described at the end of Section \ref{par:GS} (other strategies lead in this case to similar results) and $D_{max}$ was set equal to 20. The time clustering provided by the greedy ICL algorithm can be observed in Figure \ref{fig:MainFIgure2}. On the left hand side, the aggregated interactions for each quarter-hour during the first day are reported. On the right hand side, interactions taking place into those time intervals assigned to the same time cluster have the same form/color. Two important things should be noticed:
\begin{itemize}
 \item[1.] The obtained clustering seems meaningful: the three time intervals with the highest interactions level are placed in the same cluster (blue), apart from all the others. More in general, each cluster is associated to a certain intensity level, so time intervals in the same cluster, not necessarily adjacent, share the same global interactivity pattern.
 \item[2.] There are not constraints on the number of abruptly changes connected with these five time clusters. In other words, time clusters do not need to be adjacent and this is the real difference between the approach considered in this paper (time clustering) and a pure segmentation one.
\end{itemize}

\section{Conclusion}\label{sec:Conc}
We proposed a non-stationary extension of the stochastic block model (SBM) allowing us to cluster nodes of a network is situations where the classical SBM fails. The approach we chose consists in partitioning the time interval over
which interactions are studied into sub-intervals of fixed length. Those intervals provide aggregated interaction counts that are increments of non homogeneous Poisson processes (NHPPs). In a SBM inspired perspective, nodes are clustered in
such a  way that  aggregated interaction  counts are  homogeneous over
clusters.  We derived  an exact  integrated classification  likelihood
(ICL) for  such a  model and  proposed to maximize it through a greedy
search  strategy.   Finally,  a  non  parametric   maximum  likelihood
estimator was developed to estimate the integrated intensity functions
of the NHPPs counting interactions between nodes. The
experiments we carried out on artificial and real world networks highlight the
capacity of the model to capture non-stationary structures in dynamic graphs.

\newpage

\appendix\label{appendix}

\section{Computational complexity}\label{sec:comp-compl}
In this section we provide details about the computational complexity of the main model presented in this paper, namely the model \textbf{A}.
Assuming that the gamma function can be computed in constant time \citep[see][]{numericalRecipes2007}, we focus on the three statistics appearing in equation \eqref{eq:L5}, namely
\begin{enumerate}
\item{
  $S_{kgu}:=\sum_{z_i=k} \sum_{z_j=g}Y_{ij}^{I_u}$, 
}
\item{$P_{kgu}:=\prod_{z_i=k} \prod_{z_j=g}Y_{ij}^{I_u}!$},
\item{$R_{kg}:=|\mathcal{A}_k||\mathcal{A}_g|$}.
\end{enumerate} 
The whole computation task consists in evaluating the increase in ICL induced by nodes exchanges and merges. Those computations involves the tree quantities listed above. The tensor $\{S_{kgu}\}_{k,g \leq K, u\leq U}$ is stored in a three dimensional array, never resized, occupying a $O(K_{max}^2U)$ memory space. Hence, at any time during the algorithm its elements can be accessed and modified in constant time. The tensor $\{P_{kgu}\}_{k,g \leq K, u\leq U}$ is handled similarly and clusters sizes (we recall that $|\mathcal{A}_k|$ corresponds to the size of cluster $\mathcal{A}_k$) are also stored in arrays. In order to evaluate the ICL changes, induced by an operation, we need to maintain aggregated interaction counts for each node: for a node $i$ we have, e.g.
\begin{equation*}
 S_{igu}:=\sum_{z_j=g}Y_{ij}^{I_u},
\end{equation*}
the number of interactions from node $i$ to cluster $\mathcal{A}_g$ inside the time interval $I_u$. Similarly
\begin{equation*}
 S_{igu}':=\sum_{z_j=g}Y_{ji}^{I_u}
\end{equation*}
denotes the number of interactions from cluster $\mathcal{A}_g$ to node $i$ inside the time interval $I_u$. Other related quantities are considered.
These structures occupy a memory space of $O(N^2 U)$. 

\paragraph*{\textbf{Exchanges}} In order to evaluate the ICL increase induced by the switch of a node (say $i$) from cluster $\mathcal{A}_{k'}$ to cluster $\mathcal{A}_{l}$, we perform the following operations:
\begin{itemize}
\item $S_{k'gu}$ (respectively $S_{gk'u}$) is reduced by $S_{igu}$ ($S_{igu}'$) and $S_{lgu}$ ($S_{glu}$) is increased by the same amount;
\item $P_{k'gu}$ (respectively $P_{g'ku}$) is reduced by $P_{igu}$ ($P_{igu}'$) and $P_{lgu}$ ($P_{glu}$) is increased by the same amount;
\item $\mathcal{A}_{k'}$ ($\mathcal{A}_l$) is reduced (increased) by one.
\end{itemize}  
Although these operations are in constant time, they are involved in a sum with $(KU)$ elements (this can be seen in equation \eqref{eq:switch}), so that the total cost of the test is $O(KU)$. Since  node $i$ can be switched to $K-1$ remaining clusters and the graph has $N$ nodes, the cost of a \emph{full} exchange routine is $O(NK^2U)$.

\begin{Remark}
When a node is actually switched from its cluster to another one, all data structures are updated but the update cost is dominated by the cost of the testing phase described above. 
\end{Remark}
 
Notice that we have evaluated the total cost of one full exchange routine, i.e., in the case where all nodes are considered once. Reductions in the number of clusters (very likely to be induced by exchanges in case $K_{max}$ is high) are not taken into account.

\paragraph*{\textbf{Merges}}
The entire merge routine, consisting in a test phase and an actual merge, has a computational cost that is dominated by the cost of exchanges. Consider a cluster $\mathcal{A}_{k'}$. We first look for the cluster (say $\mathcal{A}_l$) leading to the best merge (highest increase in the ICL) with $\mathcal{A}_{k'}$. This operation has a cost of $O(K^2U)$: for each $\mathcal{A}_l$ the evaluation of the increase in ICL has a cost of $O(KU)$ (see equation \eqref{eq:merge}) and $l$ can take $K-1$ possible values.
Since we look for the best merge for all $k' \in \{1,\dots, K \}$ the computational cost for a merge of two nodes clusters is $O(K^3U)$, where we recall that $D\leq N$.   

\paragraph*{\textbf{Total cost}}
The worst case complexity for one iteration of the algorithm, with each node considered once, is $O(NK^2U)$. However, it is difficult to evaluate the actual complexity of the whole algorithm for two reasons.
Firstly, we have no way to estimate the number of exchanges needed in the exchange phase. Secondly, nodes exchanges are very likely to reduce the number of clusters, especially at the beginning of the algorithm, when $K_{max}$ is relatively high. Thus the individual cost of an exchange reduces very quickly leading to a vast overestimation of its cost using the proposed bounds. A detailed evaluation of the behaviour of the proposed algorithm, although outside the scope of the this paper, would be necessary to assess its use on large data sets.

\section*{References}

\bibliography{mybibfile}

\begin{thebibliography}{26}
\expandafter\ifx\csname natexlab\endcsname\relax\def\natexlab#1{#1}\fi
\expandafter\ifx\csname url\endcsname\relax
  \def\url#1{\texttt{#1}}\fi
\expandafter\ifx\csname urlprefix\endcsname\relax\def\urlprefix{URL }\fi

\bibitem[{Biernacki et~al.(2000)Biernacki, Celeux, and
  Govaert}]{biernacki2000assessing}
Biernacki, C., Celeux, G., Govaert, G., 2000. Assessing a mixture model for
  clustering with the integrated completed likelihood. Pattern Analysis and
  Machine Intelligence, IEEE Transactions on 22~(7), 719--725.

\bibitem[{Blondel et~al.(2008)Blondel, loup Guillaume, Lambiotte, and
  Lefebvre}]{Blondel08fastunfolding}
Blondel, V.~D., loup Guillaume, J., Lambiotte, R., Lefebvre, E., 2008. Fast
  unfolding of communities in large networks.

\bibitem[{C\^ome and Latouche(2015)}]{Come_Latouche15}
C\^ome, E., Latouche, P., 2015. Model selection and clustering in stochastic
  block models based on the exact integrated complete data likelihood.
  Statistical Modelling 15~(6), 564--589.

\bibitem[{Corneli et~al.(2015)Corneli, Latouche, and Rossi}]{corneli_asonam}
Corneli, M., Latouche, P., Rossi, F., Aug. 2015. {Modelling time evolving
  interactions in networks through a non stationary extension of stochastic
  block models}. In: Pei, J., Silvestri, F., Tang, J. (Eds.), {International
  Conference on Advances in Social Networks Analysis and Mining ASONAM 2015}.
  {IEEE/ACM}, {ACM}, Paris, France, pp. 1590--1591.
\newline\urlprefix\url{https://hal.archives-ouvertes.fr/hal-01263540}

\bibitem[{Dubois et~al.(2013)Dubois, Butts, and Smyth}]{proceedingsdubois2013}
Dubois, C., Butts, C., Smyth, P., 2013. Stochastic blockmodelling of relational
  event dynamics. In: International Conference on Artificial Intelligence and
  Statistics. Vol. 31 of the Journal of Machine Learning Research Proceedings.
  pp. 238--246.

\bibitem[{Fortunato(2010)}]{FortunatoSurveyGraphs2010}
Fortunato, S., 2010. Community detection in graphs. Physics Reports 486~(3-5),
  75 -- 174.

\bibitem[{Goldenberg et~al.(2009)Goldenberg, Zheng, Fienberg, and
  Airoldi}]{Goldenberg09}
Goldenberg, A., Zheng, X., Fienberg, S.~E., Airoldi, E.~M., 2009. A survey of
  statistical network models. Machine Learning 2~(2), 129--133.

\bibitem[{Guigour{\`e}s et~al.(2012)Guigour{\`e}s, Boull{\'e}, and
  Rossi}]{Rossi12}
Guigour{\`e}s, R., Boull{\'e}, M., Rossi, F., 12 2012. A triclustering approach
  for time evolving graphs. In: Co-clustering and Applications, IEEE 12th
  International Conference on Data Mining Workshops (ICDMW 2012). Brussels,
  Belgium, pp. 115--122.

\bibitem[{Guigour{\`e}s et~al.(2015)Guigour{\`e}s, Boull{\'e}, and
  Rossi}]{GuigouresEtAl2015}
Guigour{\`e}s, R., Boull{\'e}, M., Rossi, F., 2015. Discovering patterns in
  time-varying graphs: a triclustering approach. Advances in Data Analysis and
  Classification, 1--28.
\newline\urlprefix\url{http://dx.doi.org/10.1007/s11634-015-0218-6}

\bibitem[{Holland et~al.(1983)Holland, Laskey, and Leinhardt}]{Hollands83}
Holland, P., Laskey, K., Leinhardt, S., 1983. Stochastic blockmodels: first
  steps. Social Networks 5, 109--137.

\bibitem[{Isella et~al.(2011)Isella, Stehlé, Barrat, Cattuto, Pinton, and {Van
  den Broeck}}]{Isella:2011qo}
Isella, L., Stehlé, J., Barrat, A., Cattuto, C., Pinton, J., {Van den Broeck},
  W., 2011. What's in a crowd? analysis of face-to-face behavioral networks.
  Journal of Theoretical Biology 271~(1), 166--180.

\bibitem[{Leemis(1991)}]{Leemis91}
Leemis, L.~M., 1991. Nonparametric estimation of the cumulative intensity
  function for a nonhomogeneous poisson process. Management Science 37~(7),
  886--900.
\newline\urlprefix\url{http://www.jstor.org/stable/2632541}

\bibitem[{Lorrain and White(1971)}]{White1971}
Lorrain, F., White, H., 1971. {Structural equivalence of individuals in social
  networks}. Journal of Mathematical Sociology 1~(49-80).

\bibitem[{{Matias} et~al.(2015){Matias}, {Rebafka}, and
  {Villers}}]{Matias_Poisson}
{Matias}, C., {Rebafka}, T., {Villers}, F., Dec. 2015. {Estimation and
  clustering in a semiparametric Poisson process stochastic block model for
  longitudinal networks}. ArXiv e-prints.

\bibitem[{Noack and Rotta(2008)}]{NoakRotta}
Noack, A., Rotta, R., 2008. Multi-level algorithms for modularity clustering.
  CoRR abs/0812.4073.
\newline\urlprefix\url{http://arxiv.org/abs/0812.4073}

\bibitem[{Nouedoui and Latouche(2013)}]{nouedoui2013}
Nouedoui, L., Latouche, P., 2013. {Bayesian non parametric inference of
  discrete valued networks}. In: {21-th European Symposium on Artificial Neural
  Networks, Computational Intelligence and Machine Learning (ESANN 2013)}.
  Bruges, Belgium, pp. 291--296.

\bibitem[{Press et~al.(2007)Press, Teukolsky, Vetterling, and
  Flannery}]{numericalRecipes2007}
Press, W.~H., Teukolsky, S.~A., Vetterling, W.~T., Flannery, B.~P., 2007.
  {Numerical Recipes 3rd Edition: The Art of Scientific Computing}, 3rd
  Edition. Cambridge University Press.

\bibitem[{Rand(1971)}]{rand1971objective}
Rand, W.~M., 1971. Objective criteria for the evaluation of clustering methods.
  Journal of the American Statistical association 66~(336), 846--850.

\bibitem[{Schaeffer(2007)}]{Schaeffer:COSREV2007}
Schaeffer, S.~E., August 2007. Graph clustering. Computer Science Review 1~(1),
  27--64.

\bibitem[{Wang and Wong(1987)}]{wang1987}
Wang, Y., Wong, G., 1987. Stochastic blockmodels for directed graphs. Journal
  of the American Statistical Association 82, 8--19.

\bibitem[{Wasserman and Faust(1994)}]{wasserman1994social}
Wasserman, S., Faust, K., 1994. Social network analysis: Methods and
  applications. Vol. 506. Cambridge University Press.

\bibitem[{White et~al.(1976)White, Boorman, and Breiger}]{White1976}
White, H.~C., Boorman, S., Breiger, R., 1976. Social structure from multiple
  networks: I. blockmodels of roles and positions. Am. J. of Sociology 81~(4),
  730--80.

\bibitem[{Wyse et~al.(2014)Wyse, Friel, and Latouche}]{wyse2014inferring}
Wyse, J., Friel, N., Latouche, P., 2014. Inferring structure in bipartite
  networks using the latent block model and exact icl. arXiv preprint
  arXiv:1404.2911.

\bibitem[{Xing et~al.(2010)Xing, Fu, and Song}]{xing2010}
Xing, E.~P., Fu, W., Song, L., 06 2010. A state-space mixed membership
  blockmodel for dynamic network tomography. Ann. Appl. Stat. 4~(2), 535--566.

\bibitem[{Xu and Hero~III(2013)}]{xu2013dynamic}
Xu, K.~S., Hero~III, A.~O., 2013. Dynamic stochastic blockmodels: Statistical
  models for time-evolving networks. In: Social Computing, Behavioral-Cultural
  Modeling and Prediction. Springer, pp. 201--210.

\bibitem[{Yang et~al.(2011)Yang, Chi, Zhu, Gong, and Jin}]{yang2011detecting}
Yang, T., Chi, Y., Zhu, S., Gong, Y., Jin, R., 2011. Detecting communities and
  their evolutions in dynamic social networks—a bayesian approach. Machine
  learning 82~(2), 157--189.

\end{thebibliography}

\end{document}